%% file: main.tex
\ifdefined\pdfoutput\pdfoutput=1\fi
\documentclass[pdflatex,sn-mathphys-num]{lucid}
\usepackage[table]{xcolor}
\usepackage{graphicx}
\usepackage{amsmath,amssymb,amsthm}
\usepackage{textcomp}
\usepackage{booktabs,array,colortbl,tabularx}
\usepackage[skins]{tcolorbox}
\usepackage{pdflscape}
\input{leap_style.tex}

\begin{document}

\title[Look Before You LEAP]{Look Before You LEAP:\\Mitigating the Imitation Gap in Privileged Vision On-Policy Distillation with Verifiable Rewards}

\author[1,2]{\fnm{Pengyu} \sur{Dai}}
\author[2,1]{\fnm{Hongruixuan} \sur{Chen}}
\author[1]{\fnm{Junjue} \sur{Wang}}
\author[2,1]{\fnm{Jian} \sur{Song}}
\author[2]{\fnm{Yafei} \sur{Ou}}
\author[1]{\fnm{Yuzeng} \sur{Chen}}
\author[2]{\fnm{Junshi} \sur{Xia}}
\author*[1,2]{\fnm{Naoto} \sur{Yokoya}}\email{yokoya@k.u-tokyo.ac.jp}

\affil[1]{\orgdiv{Graduate School of Frontier Sciences}, \orgname{The University of Tokyo}, \orgaddress{\city{Kashiwa}, \postcode{277-8561}, \state{Chiba}, \country{Japan}}}
\affil[2]{\orgdiv{RIKEN Center for Advanced Intelligence Project (AIP)}, \orgname{RIKEN}, \orgaddress{\city{Ch\=u\=o-ku}, \postcode{103-0027}, \state{Tokyo}, \country{Japan}}}

\logo[height=10.4mm]{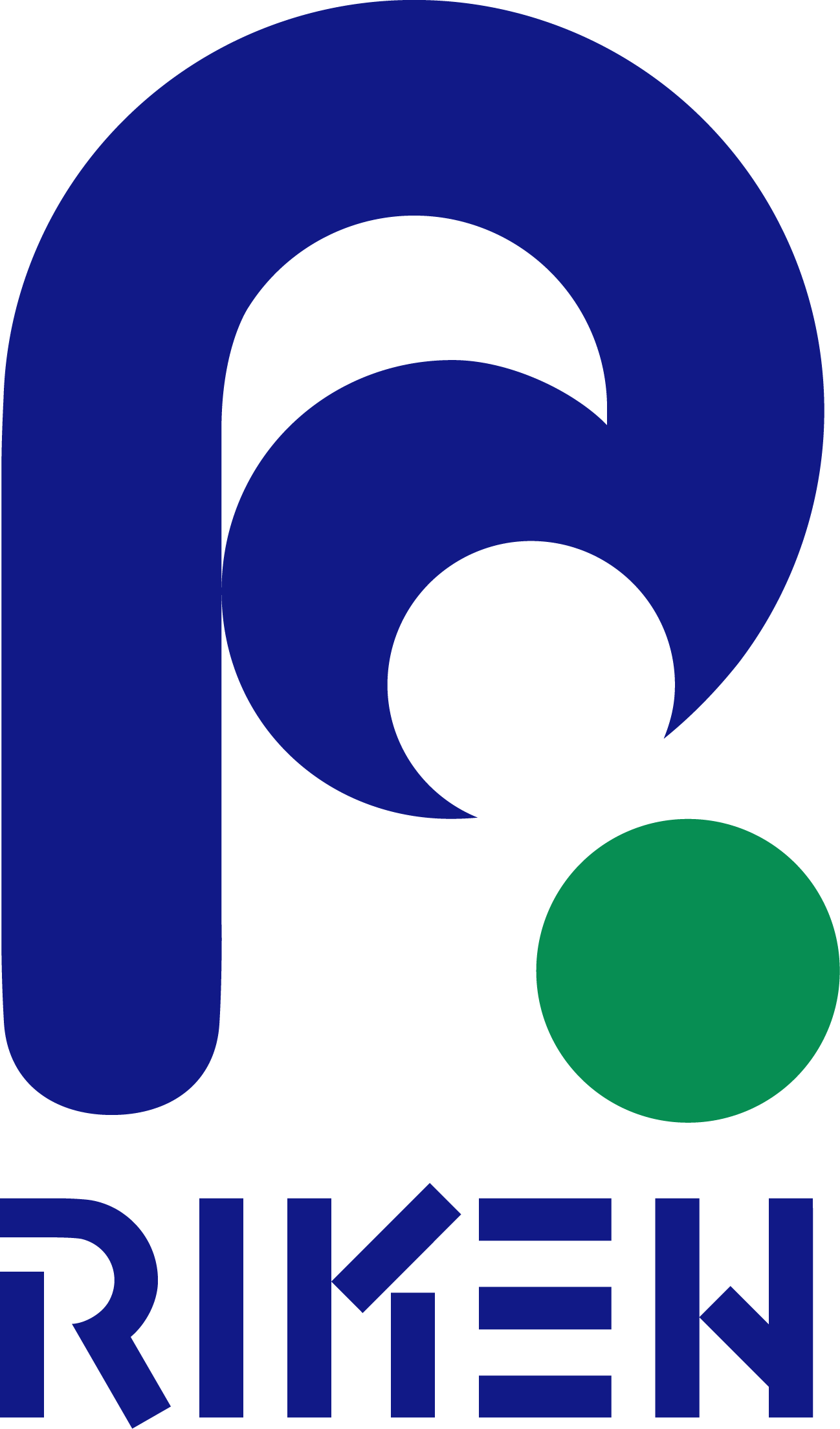}
\logo[height=7.6mm]{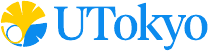}

\input{sec/0_abstract.tex}

\maketitle

\input{sec/1_introduction.tex}
\input{sec/2_related_work.tex}
\input{sec/3_method.tex}
\input{sec/4_1_experimental_setup.tex}
\input{sec/4_2_main_results.tex}
\input{sec/4_3_ablations.tex}
\input{sec/4_4_evidence_acquisition.tex}
\input{sec/4_5_remote_sensing.tex}
\input{sec/5_conclusion.tex}
\input{sec/6_llm_usage.tex}

\renewcommand\bibsection{\section*{\pdfbookmark[1]{\refname}{references}\refname}}
\bibliography{main}

\clearpage
\appendix
\setcounter{table}{0}
\renewcommand{\thetable}{S\arabic{table}}
\renewcommand{\theHtable}{S\arabic{table}}
\setcounter{figure}{0}
\renewcommand{\thefigure}{S\arabic{figure}}
\renewcommand{\theHfigure}{S\arabic{figure}}
\section*{\pdfbookmark[1]{Appendix}{appendix}Appendix}
\supptoc
\clearpage

\input{sec/C_analysis.tex}
\input{sec/A_additional_results.tex}
\input{sec/A0_evaluation_protocol.tex}
\input{sec/B_remote_sensing_protocol.tex}
\input{sec/D_implementation.tex}
\input{sec/E_intervals.tex}
\input{sec/G_prompts.tex}
\input{sec/H_case_studies.tex}

\end{document}

%% file: leap_style.tex
\definecolor{leapink}{HTML}{994455}
\definecolor{leapdark}{HTML}{71333F}
\definecolor{leapwash}{HTML}{F6E8EC}
\definecolor{leapavg}{HTML}{F0F3F7}
\definecolor{tabgray}{HTML}{5F6C79}
\definecolor{tabrule}{HTML}{243445}
\definecolor{tabinner}{HTML}{A3AEBA}
\definecolor{tabfaint}{HTML}{9AA6B2}
\newcommand{\tabsetup}[3]{\fontsize{#1}{#2}\selectfont\setlength{\tabcolsep}{#3}\renewcommand{\arraystretch}{1.0}}
\newcommand{\tabgroup}[2]{\multicolumn{#1}{l}{\textcolor{tabgray}{\textit{#2}}}\\}
\newcommand{\tabgroupavg}[2]{\multicolumn{#1}{l}{\textcolor{tabgray}{\textit{#2}}} & \\}
\newcolumntype{A}{>{\columncolor{leapavg}}c}
\newcolumntype{C}{>{\centering\arraybackslash}X}
\newcolumntype{G}{>{\columncolor{leapavg}\centering\arraybackslash}X}
\newcolumntype{L}{>{\raggedright\arraybackslash}X}
\newcommand{\leaprow}{\rowcolor{leapwash}\leavevmode\color{leapdark}}
\newcommand{\tabna}{\textcolor{tabfaint}{--}}
\let\tabBTtoprule\toprule
\let\tabBTmidrule\midrule
\let\tabBTcmidrule\cmidrule
\let\tabBTbottomrule\bottomrule
\renewcommand{\toprule}{\arrayrulecolor{tabrule}\tabBTtoprule}
\renewcommand{\midrule}{\arrayrulecolor{tabinner}\tabBTmidrule}
\renewcommand{\cmidrule}{\arrayrulecolor{tabinner}\tabBTcmidrule}
\renewcommand{\bottomrule}{\arrayrulecolor{tabrule}\tabBTbottomrule}

\theoremstyle{definition}
\newtheorem{leaplemma}{Lemma}
\newtheorem{leapproposition}[leaplemma]{Proposition}
\newtheorem{leapcorollary}[leaplemma]{Corollary}
\makeatletter
\newenvironment{leapitems}{%
  \begin{list}{}{%
    \setlength{\labelwidth}{1.6em}%
    \setlength{\labelsep}{0.35em}%
    \setlength{\leftmargin}{1.95em}%
    \setlength{\itemindent}{0pt}%
    \setlength{\listparindent}{0pt}%
    \setlength{\topsep}{2pt}%
    \setlength{\partopsep}{0pt}%
    \setlength{\parsep}{0pt}%
    \setlength{\itemsep}{3pt}%
    \@beginparpenalty=\@M
  }%
}{\end{list}}
\renewcommand\paragraph{\@startsection{paragraph}{4}{\z@}%
  {5\p@ \@plus 2\p@ \@minus \p@}{-.6em}{\normalfont\normalsize\bfseries}}
\makeatother
\newcommand{\leapstep}[1]{\par\smallskip\noindent(#1)~}

\newcounter{takeaway}
\newcommand{\boxlabel}[1]{{\normalfont\sffamily\bfseries\fontsize{8.6}{11}\selectfont\color{lcdaccent}\MakeUppercase{#1}}}
\newcommand{\takeaway}[1]{\refstepcounter{takeaway}%
  \begin{tcolorbox}[enhanced, width=\linewidth, colback=lcdpanel, colframe=lcdpanel, boxrule=0pt, arc=3.2mm,
    left=4.8mm, right=4.8mm, top=3.6mm, bottom=3.6mm, before skip=14pt, after skip=14pt]
  \boxlabel{Takeaway \thetakeaway}\par\vspace{2pt}#1\end{tcolorbox}}

\newtcolorbox{promptbox}[1]{enhanced, width=\linewidth, colback=black!3!white, colframe=lcdrule,
  boxrule=0.5pt, arc=2mm, left=4mm, right=4mm, top=3mm, bottom=3mm, before skip=12pt, after skip=12pt,
  fontupper=\small\raggedright\frenchspacing,
  before upper={{\normalfont\sffamily\bfseries\fontsize{9.6}{12}\selectfont\color{lcdaccent}#1}\par\vspace{3pt}}}
\newcommand{\ph}[1]{{\rmfamily\itshape\{#1\}}}

\makeatletter
\newcommand{\supptoc}{%
  \addtocontents{toc}{\protect\setcounter{tocdepth}{2}}%
  \edef\supptoc@depth{\the\c@tocdepth}%
  \begingroup
    \setcounter{tocdepth}{-1}%
    \parskip\z@ \parindent\z@ \hypersetup{linkcolor=black}%
    \def\numberline##1{\hb@xt@\@tempdima{{\color{lcdaccent}##1}\hfil}}%
    \def\l@section##1##2{\ifnum\c@tocdepth>\z@
      \addvspace{7\p@}\setlength\@tempdima{1.5em}%
      {\normalfont\bfseries\rightskip\@pnumwidth \parfillskip-\@pnumwidth \leavevmode
       ##1\nobreak\hfil\nobreak\hb@xt@\@pnumwidth{\hss\normalfont ##2}\par}\fi}%
    \def\l@subsection{\@dottedtocline{2}{1.5em}{2.4em}}%
    \@starttoc{toc}%
  \endgroup
  \setcounter{tocdepth}{\supptoc@depth}}
\makeatother

%% file: sec/0_abstract.tex
\begin{abstract}
Multimodal Large Language Models (MLLMs) struggle with fine-grained visual understanding, where decisive evidence is easily overlooked. Recent privileged vision on-policy distillation (OPD) trains a full-image student to imitate a teacher given privileged visual evidence (e.g., an enlarged ground-truth crop view). However, token-level imitation transfers the teacher's confident answering behavior instead of perceptual access. This induces a two-level imitation gap: an end-to-end yes-bias, in which the student affirms false visual claims more often than its base model, and behavior-level suppression of native tool use.
To let the student acquire the teacher's view from its own image instead of imitating across the gap, we propose LEAP (Learning to Equalize Access to the Privileged View), shifting privileged distillation from passive answer imitation to active evidence acquisition. LEAP factorizes one policy into trigger, tool-call, and answer tokens, each with its own supervision: the student's direct-answer failures calibrate when to look, overlap with the evidence region rewards where it looks, and contrastive privileged distillation teaches what to answer. Across six fine-grained benchmarks, LEAP improves over Vision-OPD by 5.49 points at 4B and 3.49 at 9B and on average outperforms prior privileged distillation, agentic ``Thinking with Images'' methods, closed-source models, and larger open-source models. It mitigates the imitation gap by cutting false affirmations below the released methods, preserving 87\% native zoom calls, matching ground-truth evidence views on ZoomBench with self-acquired views, and transferring zero-shot to remote sensing. Code and models will be released at \mbox{\url{https://github.com/Da1daidaidai/LeapOPD}}.
\end{abstract}

%% file: sec/1_introduction.tex
\section{Introduction}
\label{sec:introduction}

\input{fig/fig2_case.tex}

\par Fine-grained visual understanding remains challenging for Multimodal Large Language Models (MLLMs). The decisive evidence, such as the letters on a distant sign or the color of a truck's cab (Fig.~\ref{fig:case}a), often occupies only a tiny fraction of a high-resolution image and is easily overlooked~\citep{wu2024vstar,wang2025hrbench,zhang2025mmerealworld,wei2026zwz}. Notably, the bottleneck is often perceptual access rather than recognition: when the evidence region is cropped and enlarged, the same model usually answers correctly~\citep{yuan2026visionopd,zhang2025vicrop}.

\input{fig/fig1_teaser.tex}

\par Privileged vision on-policy distillation (OPD) turns this observation into a training signal~\citep{yuan2026visionopd,zhang2026vad,qiu2026opdaha,aniri2026opdv}. A full-image student is trained, at every token of its own rollouts, to match a crop-conditioned teacher, typically a frozen copy of itself that receives the evidence crop as privileged input. Unlike ``Thinking with Images'' (TwI) agents~\citep{su2025pixelreasoner,hong2026deepeyesv2,chng2025sensenovamars,zheng2026deepeyes,zhang2025thyme} trained with sparse outcome rewards~\citep{yue2025rlvr}, OPD provides dense token-level supervision. With only a few thousand examples, Vision-OPD outperforms outcome-reward GRPO trained on the same data (76.75 against 71.72 on the six-benchmark average at 4B, Table~\ref{tab:main}). The student is thereby expected to internalize zooming and to answer in a single pass without tool calls~\citep{yuan2026visionopd,wei2026zwz,wei2026zwz2}. This expectation raises our central question:

\emph{Does privileged vision distillation internalize fine-grained perception or mainly imitate the teacher's answering behavior?}

\par Our analysis and experiments show that passive OPD mainly imitates answering behavior (Fig.~\ref{fig:teaser}): the student inherits the teacher's confidence without its perceptual access. Because the teacher is given the evidence, it answers confidently and never searches for it, and a student trained on its outputs learns to do the same without the evidence. At the answer level, students from four released methods (Vision-OPD, VAD, OPD-Aha, and OPD-V) exhibit a yes-bias: they answer ``yes'' to false visual claims 14 to 28 points more often than their base model and abstain far less when the evidence is masked. At the behavior level, tool use is suppressed. The base model invokes its native zoom tool on every question, whereas distilled students do so on as few as 7\% and lose accuracy when they decide for themselves, although a forced call still localizes the evidence.

\par We attribute this two-level \emph{imitation gap}~\citep{weihs2021advisor} to the information asymmetry between teacher and student. Refining the distillation target, as prior remedies do~\citep{zhang2026vad,qiu2026opdaha,aniri2026opdv}, does not remove the gap, and in our controlled comparison a contrastive target suppresses tool calls even further (Sec.~\ref{sec:leap_gap}). A student that cannot resolve the evidence can at best match the teacher's average prediction over inputs it cannot distinguish, and the remaining error, the \emph{observation residual}, cannot be reduced by better fitting. Moreover, the teacher's outputs never demonstrate how to find the evidence~\citep{walsman2023impossibly}. In vision, however, this asymmetry is removable. Unlike privileged text~\citep{zhang2026caopd,peng2026adopsd}, the teacher's crop is taken from the student's own image, so the teacher sees no additional pixels; the same region simply occupies far more of its visual tokens. One zoom on the student's image recovers the teacher's view, and the crop box verifies that it did without extra annotation. Instead of imitating across the gap, the student should first close it.

\par We propose \textbf{LEAP} (\emph{Learning to Equalize Access to the Privileged View}), shifting OPD from passive answer imitation to active evidence acquisition.
Doing so requires three decisions: \emph{when} the full image is insufficient, \emph{where} the evidence lies, and \emph{what} to answer once it is in view. LEAP casts them as three consecutive token segments of a single policy (a trigger token, tool-call tokens that specify a region, and answer tokens) and supervises each with a signal that the privileged setup already provides. The teacher's distribution supervises \emph{what to answer} through OPD-Aha's contrastive target~\citep{qiu2026opdaha}. The bounding box that defines the teacher's crop supervises \emph{where to look}. Unlike Vision-OPD~\citep{yuan2026visionopd} and OPD-Aha, which use the evidence region only to crop the teacher's view and mark the training image, LEAP also uses it to reward clean-image region selection. The student's own failures when it answers without the view indicate \emph{when to look}. At inference, the policy looks before it leaps: in our \emph{test-time acquisition procedure}, it decides whether to zoom, acquires a view, and answers from both views (Fig.~\ref{fig:framework}).

\input{fig/fig3_framework.tex}

Through our experiments, LEAP-4B and LEAP-9B raise the six-benchmark average over naive Vision-OPD by 5.49 and 3.49 points and surpass other passive OPD students, TwI agents, and larger open-source models and closed-source models (Fig.~\ref{fig:case}, Table~\ref{tab:main}).
LEAP-4B also mitigates both levels of the imitation gap: it cuts false affirmations below the four released methods without losing net claim verification, keeps 87\% native zoom calls, and ranks its own direct-answer failures above chance (Tables~\ref{tab:gap}, \ref{tab:bias} and~\ref{tab:bat}). Test-time cropping alone does not explain these gains: zooming lifts the base model but hurts Vision-OPD-4B, while LEAP-4B's self-acquired views match ground-truth evidence views on ZoomBench (Fig.~\ref{fig:gain-source}c).
Meanwhile, LEAP-9B transfers zero-shot to remote sensing and matches the domain-trained RS-OPSD-8B~\citep{jiang2026rsopsd} on average (Table~\ref{tab:remote-sensing}).

Our contributions are:
\begin{itemize}
\item Diagnosing the two-level \emph{imitation gap} in OPD (end-to-end yes-bias and suppressed native tool use), traced to the observation residual and missing search (Sec.~\ref{sec:leap_gap}, Fig.~\ref{fig:teaser}).
\item Proposing \textbf{LEAP} to mitigate both the observation residual and fitting error in one factorized policy, where direct-answer failures and privileged view overlap supervise active evidence acquisition alongside contrastive answer distillation (Sec.~\ref{sec:leap}, Fig.~\ref{fig:framework}).
\item Demonstrating that LEAP lifts Avg$_6$ by $5.49$/$3.49$ points over Vision-OPD at 4B/9B, matches ground-truth views on ZoomBench, curbs yes-bias, keeps 87\% native zoom calls, and transfers zero-shot to remote sensing (Sec.~\ref{sec:experiments}).
\end{itemize}

%% file: fig/fig2_case.tex
\begin{figure}[t]
\centering
\includegraphics[width=0.66\linewidth,trim=0 28 0 0,clip]{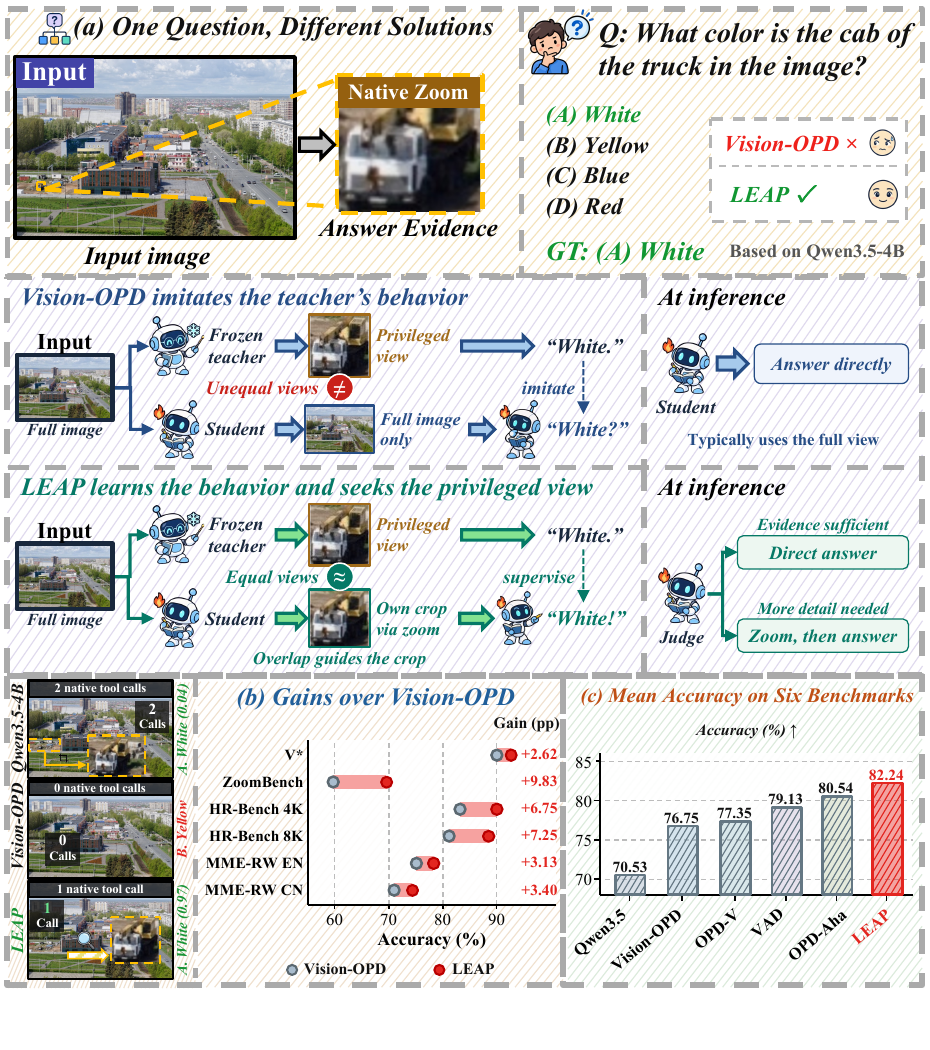}
\caption{\textbf{LEAP shifts privileged distillation from passive imitation to active evidence acquisition.} (a) Instead of imitating a crop teacher across unequal views and erring without zooming (Vision-OPD), LEAP equalizes views via overlap-guided zooming (one call, IoU 0.97), (b) gaining up to $+9.83$ points across all six benchmarks and (c) lifting the 4B mean to 82.24\% (Table~\ref{tab:main}).}
\label{fig:case}
\end{figure}

%% file: fig/fig1_teaser.tex
\begin{figure}[t]
\centering
\includegraphics[width=0.66\linewidth]{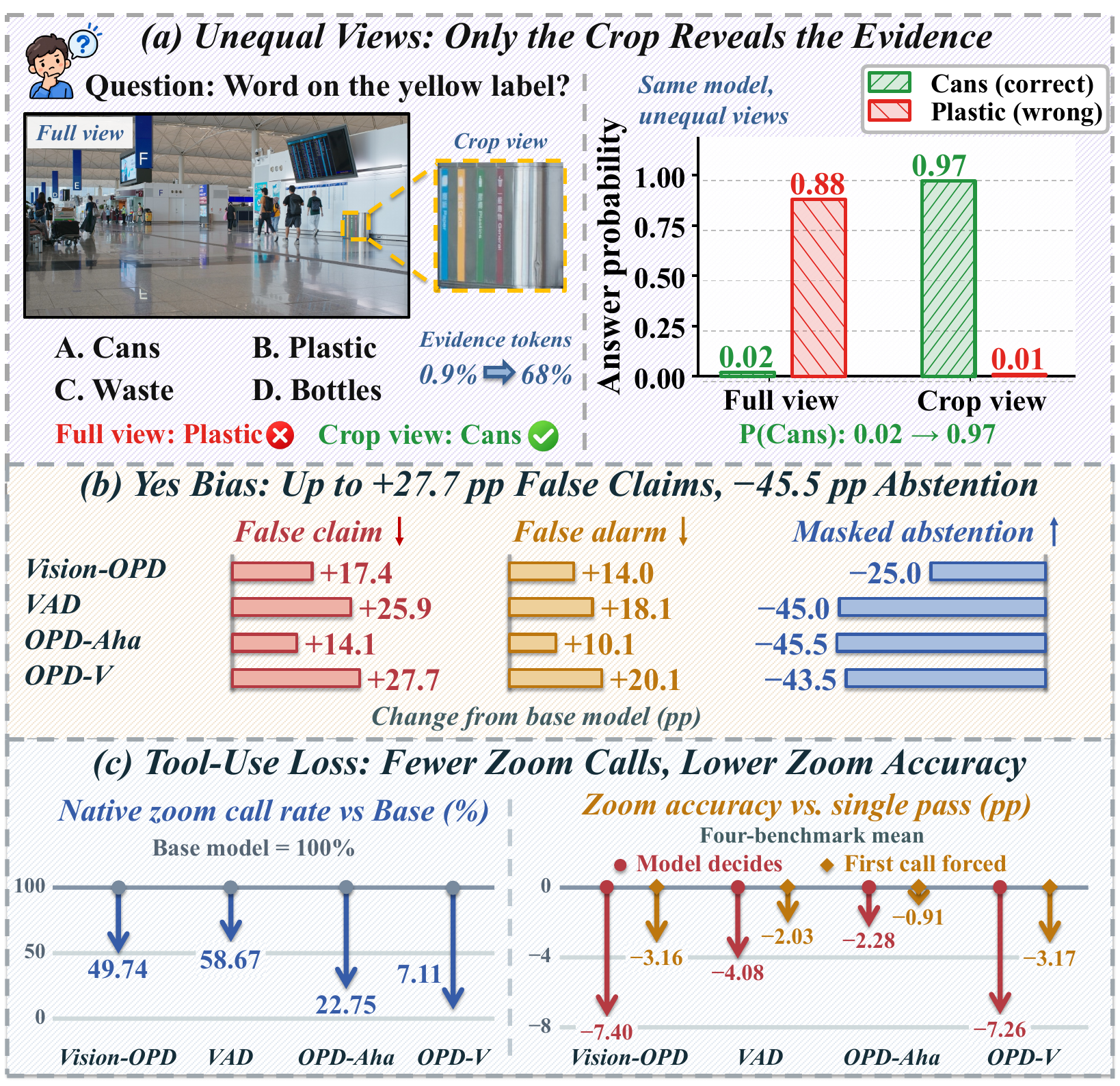}
\caption{\textbf{Imitating across unequal views induces a two-level imitation gap.} (a) Here, evidence fills 0.9\% of full-view tokens vs.\ 68\% of the crop. Matching the teacher transfers confident behavior without access: four released 4B students (b) affirm more false claims and abstain less on masked evidence, and (c) zoom less and lose accuracy even under forced calls (Tables~\ref{tab:bias}, \ref{tab:native}, \ref{tab:tool-loops}).}
\label{fig:teaser}
\end{figure}

%% file: fig/fig3_framework.tex
\begin{figure}[t]
\centering
\includegraphics[width=\linewidth]{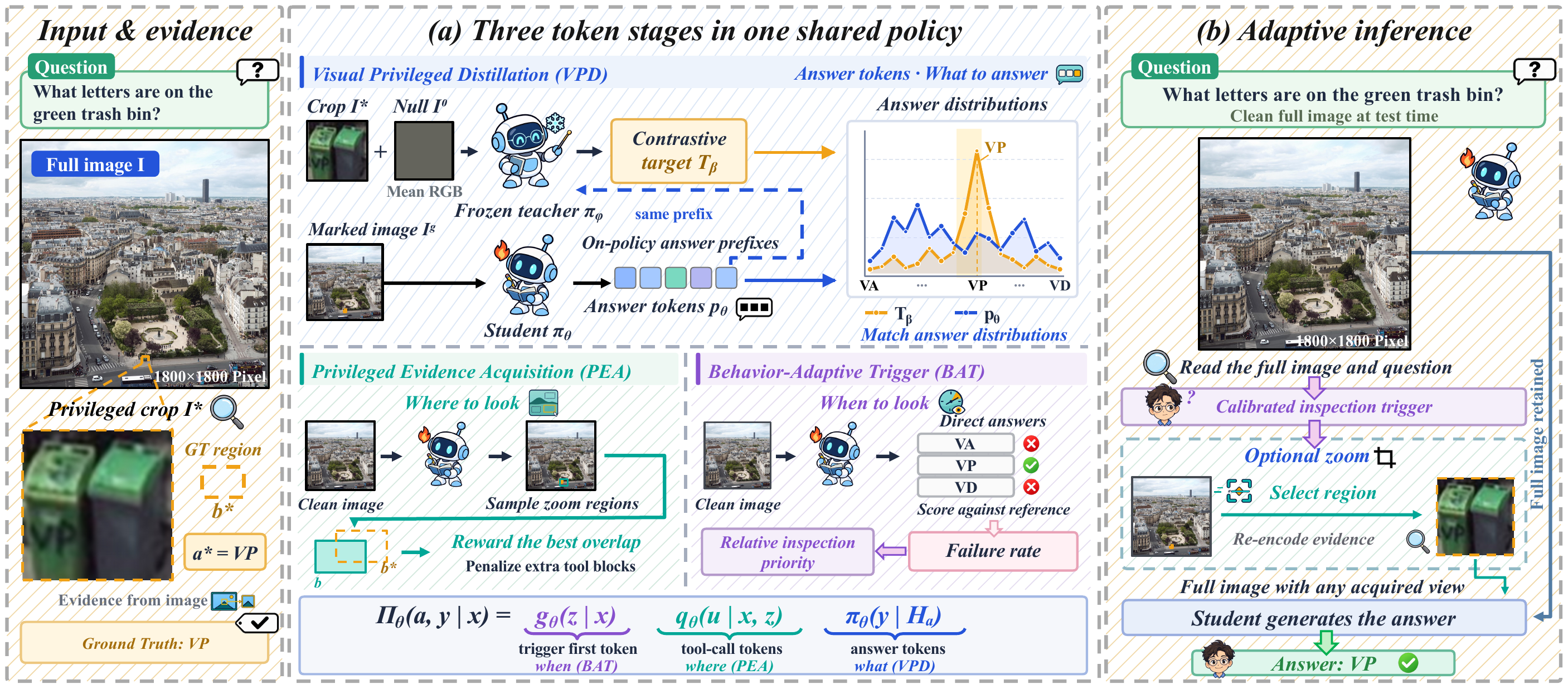}
\caption{\textbf{LEAP equalizes perceptual access by supervising three token stages in one factorized policy.} (a) In training, one policy $\Pi_\theta(a,y\mid x)$ learns \emph{what to answer} by distilling a crop teacher's contrastive target $T_t^\beta$ (VPD), \emph{where to look} by rewarding box $b$'s overlap with $b^\star$ (PEA), and \emph{when to look} by calibrating trigger $g_\theta$ from direct-answer failures $e$ (BAT). (b) At test time, it zooms when needed and answers from both views. Examples are illustrative.}
\label{fig:framework}
\end{figure}

%% file: sec/2_related_work.tex
\section{Related Work}
\label{sec:related_work}

\paragraph{Fine-grained perception with visual tools.}
\label{sec:rw_high_resolution}
\par High-resolution MLLMs first encoded images as separate tiles~\citep{li2024monkey,guo2024llavauhd} and later selected visual inputs outside the model, by pruning tokens~\citep{chen2024fastv,yang2025visionzip}, cropping attended regions~\citep{zhang2025vicrop}, or retrieving relevant crops~\citep{wang2025rap}. TwI makes region selection part of the policy: V$^{\star}$~\citep{wu2024vstar} couples the answering model with guided visual search, and later agents such as DeepEyes~\citep{zheng2026deepeyes}, Pixel Reasoner~\citep{su2025pixelreasoner}, Thyme~\citep{zhang2025thyme}, and CodeV~\citep{hou2026codev} learn interleaved reasoning and image operations through reinforcement learning. Recent studies ask when tool calls actually help, by adapting tool-use supervision to its benefit (ToolVision~\citep{mao2026toolvision}, AdaTooler-V~\citep{wang2026adatoolerv}, ToolsRL~\citep{dong2026toolsrl}) or by auditing whether tool calls drive the gains~\citep{ma2026med,wang2026illusion,kim2026diversity}. LEAP needs no outcome reward on the tool-call tokens: where to look is scored by the region defining the teacher's crop, and when to look comes from direct-answer failures.

\paragraph{Privileged vision distillation.} \label{sec:rw_privileged_distillation}
\par This line of work trains a full-image student with supervision derived from enlarged evidence crops, either as generated question--answer pairs (Zooming without Zooming, ZwZ~\citep{wei2026zwz,wei2026zwz2}) or as token-level targets from a crop-conditioned teacher on the student's own rollouts (Vision-OPD~\citep{yuan2026visionopd}). Follow-up work refines what is distilled, through target reconstruction (VAD~\citep{zhang2026vad}, OPD-Aha~\citep{qiu2026opdaha}), position selection (OPD-V~\citep{aniri2026opdv}), projection onto the student's realizable response space (FP-OPD~\citep{xue2026fpopd}), or internal recovery of visual cues (ViCuR~\citep{tian2026vicur}). Because all of these keep the student's visual input fixed, they only refine fitting on an unresolved view; LEAP instead acquires the privileged view before answering while distilling OPD-Aha's target on the answer tokens.

\paragraph{Imitation under information asymmetry.}
It is well known in imitation learning that a student may be unable to imitate a teacher with privileged information. ADVISOR~\citep{weihs2021advisor} terms the resulting discrepancy the imitation gap and down-weights imitation where the student cannot infer the teacher's action, and \citet{walsman2023impossibly} show that an ``impossibly good'' expert never demonstrates the information-gathering actions its follower needs. For language models, on-policy distillation under privileged context has been linked to overconfidence (CaOPD~\citep{zhang2026caopd}) and suppressed intermediate reasoning (AD-OPSD~\citep{peng2026adopsd}). We observe the visual counterparts of both symptoms and exploit a property these settings lack: one zoom on the student's own image reproduces the teacher's view, and the annotation that defines the crop scores that action.

%% file: sec/3_method.tex
\section{LEAP: The Imitation Gap and How to Mitigate It}
\label{sec:leap}

\paragraph{Setup.} Each training example consists of an image $I$, a question $q$, a reference answer $a^\star$, and an evidence region $b^\star$ given as a bounding box. The student $\pi_\theta$ observes the clean input $x=(I,q)$, whereas the teacher $\pi_\phi$, a frozen copy of the initial student, observes the crop $I^\star=\operatorname{Crop}(I,b^\star)$ instead of $I$.
At inference, $b^\star$ is unavailable: the student receives only $x$ and selects its own regions.
We first analyze what a student cannot learn by imitating this teacher alone (Sec.~\ref{sec:leap_gap}), and then train it to acquire the teacher's view by itself (Secs.~\ref{sec:leap_policy} and~\ref{sec:leap_training}; Fig.~\ref{fig:framework}).

\subsection{What Imitation Cannot Transfer}
\label{sec:leap_gap}

\paragraph{The teacher's advantage is perceptual access.}
Since the crop is taken from the student's image, the teacher sees no additional pixels. Its advantage is \emph{perceptual access}: over the training examples, the evidence occupies a median of 1.1\% of the student's visual tokens but 68\% of the teacher's, which alone raises the base model's forced-choice accuracy from 50.6\% to 77.2\%. Students that imitate such a teacher from the full view inherit its behavior without this access, at two levels (Fig.~\ref{fig:teaser}). At the answer level, the four released 4B methods answer ``yes'' to false visual claims more often than their base model, without verifying claims more accurately under original and reversed questions (Table~\ref{tab:bias}); on POPE~\citep{li2023pope}, their false-alarm rates increase by 7.7 to 14.3 points at 4B and by 0.8 to 10.5 at 9B. At the behavior level, they invoke their native zoom tool on only 7.1\% to 58.7\% of questions, against 100\% for the base model, and recover fewer errors with it, even though a forced call still localizes the evidence (Table~\ref{tab:native}).
Our own distillation of Qwen3.5-4B reproduces both effects. The gap appears under both targets we test, and the contrastive one suppresses tool calls further: the plain crop teacher and VPD (Sec.~\ref{sec:leap_training}) increase false affirmations by 8.6 and 12.5 points and reduce native zoom calls to 48.3\% and 13.0\%, and by training step 20 VPD's trigger ranks the student's own errors below chance (Fig.~\ref{fig:gain-source}b).

\paragraph{Answer fitting averages teacher targets.}
\par For a fixed representation of the student, better fitting cannot close this gap. Let $S$ denote a training example and $T_S$ the teacher's answer distribution on it. Let $X$ denote what the student's global encoding resolves of $S$, so that examples sharing the same $X$ are indistinguishable to the student, and let $\bar T_X=\mathbb E[T_S\mid X]$ be the teacher's target averaged over them. Following the conditional-policy analysis of ADVISOR~\citep{weihs2021advisor}, the loss of any student $p_X$ decomposes as
\begin{equation}
\mathbb E\,\mathrm{KL}(T_S\|p_X)
=\underbrace{\mathbb E\,\mathrm{KL}(T_S\|\bar T_X)}_{\text{observation residual }\mathcal G(X)}
+\underbrace{\mathbb E\,\mathrm{KL}(\bar T_X\|p_X)}_{\text{fitting error}},
\label{eq:leap_projection}
\end{equation}
where fitting teacher answers targets the second term, while $\mathcal G(X)$ remains whenever $X$ cannot distinguish examples with different teacher targets. At its optimum, the student predicts the average $\bar T_X$: behavior that does not depend on the evidence transfers intact, whereas evidence-dependent answers are averaged over the examples that share $X$.
If an acquired view $Z$ augments the observation to $(X,Z)$, then
\begin{equation}
\mathcal G(X)-\mathcal G(X,Z)
=\mathbb E\,\mathrm{KL}(\bar T_{X,Z}\|\bar T_X)\;\geq\;0,
\label{eq:leap_refinement_kl}
\end{equation}
with strict inequality whenever the view changes the averaged target, and the residual vanishes when the view is the teacher's own crop. An informative view therefore lowers the minimum imitation discrepancy; under JSD it cannot rise (Appendix~\ref{app:observation-residual}).

\paragraph{The teacher never demonstrates how to look.}
\par A view lowers the residual (Eq.~\eqref{eq:leap_refinement_kl}), but imitation cannot show how to obtain it.
Given the crop, the teacher answers directly: its demonstrations never inspect the image, and no reweighting of its labels can supply an action it never takes~\citep{walsman2023impossibly} (Appendix~\ref{app:missing-inspection}).
Passive distillation therefore leaves the trigger unsupervised, and in practice it is suppressed. Closing the gap at test time thus requires three capabilities: detecting \emph{when} the residual is nonzero, locating \emph{where} the view that reduces it lies, and fitting \emph{what} the teacher answers given that view. The teacher supervises only the last.

\takeaway{\textbf{Imitation transfers answering behavior, not perceptual access.} Students that match a crop teacher from the global view show end-to-end yes-bias and suppressed native tool calls, whereas zooming into the student's own image can equalize access.}

\subsection{Learning to Acquire Privileged Evidence}
\label{sec:leap_policy}
\par Since the crop is part of the student's own image, the student can acquire the view by itself with its native zoom tool. LEAP unifies \emph{when} to look, \emph{where} to look, and \emph{what} to answer in one autoregressive policy. The first generated token $z$ either opens a zoom call ($z{=}1$, \texttt{<tool\_\allowbreak call>}) or starts a direct answer ($z{=}0$). After a call, the tool-call tokens $u$ specify a box $b\in\mathcal B(u)$ and the tool returns the crop $Z_b$ (a call may return several regions); the model then answers $y$ from $H_a=(x,Z_b)$, the full image together with $Z_b$. Without a call, $u$ is empty and $H_a=x$.
By the chain rule, the policy over action $a=(z,u)$ and answer $y$ is
\begin{equation}
\Pi_\theta(a,y\,|\,x)=
\underbrace{g_\theta(z\,|\,x)}_{\substack{\text{first token}\\\text{\emph{when} (BAT)}}}\,
\underbrace{q_\theta(u\,|\,x,z)}_{\substack{\text{tool-call tokens}\\\text{\emph{where} (PEA)}}}\,
\underbrace{\pi_\theta(y\,|\,H_a)}_{\substack{\text{answer tokens}\\\text{\emph{what} (VPD)}}},
\label{eq:leap_policy}
\end{equation}
whose first two factors form the acquisition policy $\rho_\theta(a\mid x)$ of Fig.~\ref{fig:framework}.
The three factors share all parameters $\theta$ and differ only in the tokens they govern and in what can supervise them.
Under forward KL, the answer factor affects only the fitting error of Eq.~\eqref{eq:leap_projection}, whereas the first two also affect the residual (Appendix~\ref{app:observation-residual}). Crucially, privileged distillation already supplies two verifiable signals without extra annotation: the evidence region $b^\star$ for \emph{where to look} and the student's direct-answer failures for \emph{when to look}.

\subsection{A Supervision Signal for Each Decision}
\label{sec:leap_training}

\paragraph{What to answer (VPD).}
\label{sec:leap_vpd}
\emph{Visual Privileged Distillation} (VPD) lowers the fitting error of the answer tokens on the full image $I^g$ with $b^\star$ marked, so answer learning does not wait for the student's own search; Sec.~\ref{sec:evidence-quality} tests transfer to acquired views.
At a student prefix $h_t$, it adopts OPD-Aha's target~\citep{qiu2026opdaha}, which contrasts the teacher's predictions on the crop $I^\star$ and on its mean-RGB null view $I^0$:
\begin{equation}
\begin{gathered}
p_t^+=\pi_\phi(\cdot\mid I^\star,q,h_t),\quad
p_t^0=\pi_\phi(\cdot\mid I^0,q,h_t),\\
T_t^\beta=\operatorname{softmax}\!\left[
\log p_t^+ +\beta(\log p_t^+-\log p_t^0)\right],
\end{gathered}
\label{eq:leap_target}
\end{equation}
with $\beta=0$ recovering the crop teacher.
VPD minimizes the divergence from the student's prediction $p_t^\theta$ on $I^g$,
\begin{equation}
\mathcal L_{\mathrm{VPD}}
=\frac1{|\mathcal T|}\sum_{t\in\mathcal T}
w_t\,\mathrm{JS}_{1/2}(T_t^\beta,p_t^\theta),
\label{eq:leap_vpd}
\end{equation}
over valid generated tokens $\mathcal T$, with detached importance weights $w_t$ (Appendix~\ref{app:leap_impl}).

\paragraph{Where to look (PEA).}
\label{sec:leap_pea}
VPD does not supervise the tool-call tokens, and $b^\star$ is unavailable at test time, so the student must learn to locate the evidence on the clean image.
\emph{Privileged Evidence Acquisition} (PEA) rewards these tokens with the region that defines the teacher's crop: $\operatorname{IoU}(b,b^\star)$ is a deterministic spatial proxy for the residual reduction that needs no teacher rollouts on candidate crops; at $\operatorname{IoU}=1$ the view is the teacher's own (Appendix~\ref{app:leap_theory}).
For each question, the student samples $M$ native zoom calls $u_j\sim q_\theta(\cdot\mid x,1)$.
A usable call with valid boxes $\mathcal B(u_j)$ and $n_j$ tool blocks earns
\begin{equation}
R_j=\max_{b\in\mathcal B(u_j)}\operatorname{IoU}(b,b^\star)-\eta(n_j-1),
\label{eq:leap_reward}
\end{equation}
which favors overlap with $b^\star$ and penalizes extra tool blocks; an unusable call earns $r_{\mathrm{bad}}$.
Group-centered advantages $A_j=R_j-\frac1M\sum_kR_k$ train the tool-call tokens with a clipped policy loss $\mathcal L_{\mathrm{PEA}}$~\citep{shao2024deepseekmath}.
A reward on the final answer could also train $q_\theta$, but because the full image stays visible in $H_a$, a correct answer does not verify that the view covers the evidence (Sec.~\ref{sec:ablations}).

\paragraph{When to look (BAT).}
\label{sec:leap_bat}
A view reduces the residual only where it is nonzero, so zooming into every image is unnecessary; yet passive distillation suppresses the trigger and even inverts its priority (Table~\ref{tab:bat}).
Neither the teacher nor $b^\star$ indicates which inputs this particular student needs to inspect, but the student's own behavior does: failing to answer directly is an observable proxy for a nonzero residual.
The \emph{Behavior-Adaptive Trigger} (BAT) estimates this proxy online: for each question $x_i$, it samples $K$ direct answers ($z{=}0$) and records their failure fraction,
\begin{equation}
e_i=\frac1K\sum_{k=1}^K
\mathbf1\!\left[\operatorname{score}\big(\hat a_i^{(k)},a_i^\star\big)=0\right].
\label{eq:leap_bat_target}
\end{equation}
BAT then trains the trigger margin $m_i^\theta=\operatorname{logit}g_\theta(x_i)$, where $g_\theta(x_i)=g_\theta(z{=}1\mid x_i)$, to rank inputs by $e_i$.
It fits an offset $\delta$ so that the mean adjusted rollout probability matches the mean failure fraction, and centers the residuals $d_i=\sigma(\bar m_i+\delta)-e_i$ of the rollout margins $\bar m_i$ over $N$ questions:
\begin{equation}
c_i=d_i-\frac1N\sum_jd_j,\quad
\mathcal L_{\mathrm{BAT}}=\frac1N\sum_i\operatorname{sg}(c_i)\,m_i^\theta,
\label{eq:leap_bat_loss}
\end{equation}
where $\operatorname{sg}$ stops gradients.
Since $\sum_ic_i=0$, the update raises the trigger where the student fails more often than predicted and lowers it elsewhere, without an explicit common margin shift (Appendix~\ref{app:leap_trigger_proof}: its fixed point ranks inputs by failure rate).
BAT thus learns \emph{which} inputs to inspect and leaves \emph{how many} to a test-time threshold.

At inference, the factors of Eq.~\eqref{eq:leap_policy} run in order as our  acquisition procedure: if $g_\theta(x)$ reaches a calibrated threshold, $q_\theta$ selects views and $\pi_\theta$ answers from $(x,Z_b)$; otherwise, $\pi_\theta$ answers from $x$ (Appendix~\ref{app:leap_impl}).

%% file: sec/4_1_experimental_setup.tex
\section{Experiments}
\label{sec:experiments}

We test four claims: whether LEAP improves fine-grained perception beyond test-time cropping or extra compute (Sec.~\ref{sec:main-results}), how VPD, PEA, and BAT govern what to answer, where to look, and when to inspect (Sec.~\ref{sec:ablations}), whether self-acquired views match ground-truth evidence views and mitigate the imitation gap (Sec.~\ref{sec:evidence-quality}), and whether learned acquisition transfers zero-shot to remote sensing (Sec.~\ref{sec:remote-sensing}).

\subsection{Experimental Setup}
\label{sec:experimental-setup}

\paragraph{Benchmarks and metrics.}
We evaluate fine-grained perception on six benchmarks: V$^{\star}$~\citep{wu2024vstar}, ZoomBench~\citep{wei2026zwz}, HR-Bench 4K and 8K~\citep{wang2025hrbench}, and the English and Chinese splits of MME-RealWorld~\citep{zhang2025mmerealworld}.
We report accuracy under the official Vision-OPD protocol~\citep{yuan2026visionopd}, where Avg$_6$ is the unweighted mean before rounding (Table~\ref{tab:main}; paired 95\% intervals in Appendix~\ref{app:intervals}), and test transfer on XLRS-Bench-lite~\citep{wang2025xlrsbench}, MME-RealWorld-RS~\citep{zhang2025mmerealworld}, and LRS-VQA~\citep{luo2025lrsvqa} with a shared corrected judge (Appendix~\ref{app:remote-sensing}).

\paragraph{Baselines and evaluation protocol.}
We compare against Qwen3.5-based supervised fine-tuning (SFT), RL, and passive privileged distillation at 4B and 9B, TwI agents (DeepEyes~\citep{zheng2026deepeyes}, DeepEyesV2~\citep{hong2026deepeyesv2}, Thyme-RL~\citep{zhang2025thyme}, and SenseNova-MARS~\citep{chng2025sensenovamars}) with native tool loops and the official judge (tool-free: Table~\ref{tab:tools-off}), larger open-source models~\citep{qwen2026qwen35,zai2025glm46v,kimi2026k3,deepseek2026v41flash}, and closed-source models~\citep{google2026gemini31pro,google2026gemini38flash,anthropic2026haiku55,openai2026gpt6luna}.
Gemini 3.1 Pro, GRPO~\citep{shao2024deepseekmath}, VA-OPD~\citep{liu2026vaopd}, Decomposed OPD~\citep{yoon2026decomposedopd}, and 9B V-Zero~\citep{sun2026vzero} are from VAD~\citep{zhang2026vad}, ZwZ2 from~\citep{wei2026zwz2}; we evaluate the rest (Appendix~\ref{app:evaluation-details}).

\paragraph{Training and inference.}
LEAP trains Qwen3.5-4B and 9B~\citep{qwen2026qwen35} from a frozen same-backbone teacher on Vision-OPD-6K~\citep{yuan2026visionopd} ($\beta=4$ at 4B, $\beta=8$ at 9B), normalizing $\mathcal L_{\mathrm{train}}=\mathcal L_{\mathrm{VPD}}+\lambda_{\mathrm{BAT}}\mathcal L_{\mathrm{BAT}}+\lambda_{\mathrm{PEA}}\mathcal L_{\mathrm{PEA}}$ by rollout type (VPD on $I^g$, PEA and BAT on clean $x$; Appendix~\ref{app:leap_impl}).
At test time, $g_\theta(x)$ is thresholded per model at a fixed inspection share of 480 held-out examples.

%% file: sec/4_2_main_results.tex
\subsection{Main Results}
\label{sec:main-results}
\input{tab/table1_main.tex}
\input{tab/table_gap.tex}

In Table~\ref{tab:main}, LEAP achieves the highest Avg$_6$ at both scales (82.24\% at 4B and 82.56\% at 9B), gaining 5.49 and 3.49 points over Vision-OPD and 1.70 and 0.78 points over the strongest passive student OPD-Aha (Fig.~\ref{fig:case}).
At 4B, SFT (72.68\%) and outcome-reward GRPO (71.72\%) gain modestly over the base (70.53\%) and drop on HR-Bench, while passive distillation sharpens single-pass reading of the fixed global view.
By actively acquiring high-resolution views, LEAP delivers its largest gains on ZoomBench (69.59\% and 69.70\%; +9.83 and +6.75 points over Vision-OPD and OPD-Aha at 4B) and HR-Bench 4K/8K (90.00\%/88.50\% at 4B, 90.75\%/90.25\% at 9B).
LEAP-4B also leads 7B--8B ``Thinking with Images'' RL agents by 6.75 to 15.45 Avg$_6$ points (+15.27 over SenseNova-MARS on ZoomBench) and surpasses open-source models up to 2.8T (Qwen3.5-397B-A17B, 77.46\%; Kimi K3, 74.61\%) and closed-source models (Gemini 3.1 Pro, 78.04\%; Gemini 3.8 Flash, 80.79\%).

Test-time zooming or extra compute alone does not explain these gains.
On the four perception benchmarks (V$^\star$, ZoomBench, and HR-Bench 4K/8K), zooming lifts the base model and LEAP-4B by $4.22$ and $3.01$ points but lowers Vision-OPD-4B and VAD-4B by $7.40$ and $4.08$, leaving LEAP-4B $14.01$ and $7.82$ points ahead (Table~\ref{tab:tool-loops}).
Even forced zooming leaves every released student $0.91$ to $3.17$ points below its own single pass, with the closest (OPD-Aha, 81.83\%) still $3.36$ points behind LEAP-4B (85.19\%), showing that passive distillation damages tool use itself.
Three-sample majority voting of LEAP-4B's direct answers at a matched token budget adds only $0.20$ points ($2.81$ below LEAP-4B; Table~\ref{tab:tool-loops}, Appendix~\ref{app:intervals}).

%% file: tab/table1_main.tex
\begin{table}[t]
\centering
\caption{\textbf{Comparison on six visual benchmarks.} Accuracy (\%); \textbf{bold}/\underline{underline}: best/second best per Qwen3.5 scale.}
\label{tab:main}
\begingroup
\tabsetup{9.6}{11.9}{4.2pt}\setlength{\aboverulesep}{0.25ex}\setlength{\belowrulesep}{0.45ex}
\begin{tabularx}{\linewidth}{L c *{6}{c} A}
\toprule
Model & Size & V$^{\star}$ & ZoomBench & \shortstack{HR-Bench\\4K} & \shortstack{HR-Bench\\8K} & \shortstack{MME-RW\\EN} & \shortstack{MME-RW\\CN} & \textbf{Avg$_6$} \\
\midrule
\tabgroupavg{8}{Thinking with Images}
DeepEyes & 7B & 85.86 & 47.22 & 73.88 & 69.00 & 64.12 & 65.52 & 67.60 \\
Thyme-RL & 7B & 83.25 & 45.68 & 78.38 & 71.25 & 65.01 & 64.88 & 68.07 \\
DeepEyesV2 & 7B & 83.77 & 43.31 & 74.25 & 71.38 & 63.78 & 64.24 & 66.79 \\
SenseNova-MARS & 8B & 92.67 & 54.32 & 84.62 & 82.00 & 69.22 & 70.10 & 75.49 \\
\midrule
\tabgroupavg{8}{Closed-source models}
Gemini 3.1 Pro & \tabna & 88.48 & 62.01 & 88.12 & 84.50 & 73.32 & 71.81 & 78.04 \\
Gemini 3.8 Flash & \tabna & 91.62 & 62.60 & 92.25 & 88.75 & 76.05 & 73.47 & 80.79 \\
Claude Haiku 5.5 & \tabna & 87.43 & 58.46 & 83.62 & 78.50 & 72.28 & 70.32 & 75.10 \\
GPT-6-luna & \tabna & 87.43 & 60.24 & 89.62 & 63.62 & 72.53 & 70.66 & 74.02 \\
\midrule
\tabgroupavg{8}{Large Open-source models}
Qwen3.5-397B-A17B & 397B & 87.43 & 58.82 & 88.62 & 85.38 & 74.65 & 69.83 & 77.46 \\
GLM-4.6V & 106B & 82.72 & 48.40 & 79.25 & 74.62 & 67.64 & 64.37 & 69.50 \\
Kimi K3 & 2.8T & 87.43 & 58.70 & 82.75 & 79.12 & 70.14 & 69.53 & 74.61 \\
DeepSeek v4.1 Flash & 284B & 79.06 & 48.05 & 77.75 & 72.12 & 66.64 & 60.71 & 67.39 \\
\midrule
\tabgroupavg{8}{Qwen3.5-based models (4B)}
Qwen3.5 & 4B & 81.15 & 48.76 & 85.62 & 80.25 & 63.85 & 63.55 & 70.53 \\
SFT & 4B & 86.39 & 57.51 & 79.12 & 75.75 & 71.41 & 65.86 & 72.68 \\
ZwZ2 & 4B & 89.00 & 58.22 & \underline{88.62} & 83.63 & 75.48 & 73.89 & 78.14 \\
GRPO & 4B & 81.68 & 56.57 & 76.75 & 75.62 & 71.10 & 68.58 & 71.72 \\
VA-OPD & 4B & 90.05 & 58.11 & 81.88 & 78.75 & 73.29 & 67.42 & 74.92 \\
V-Zero & 4B & 76.96 & 48.40 & 84.88 & 80.75 & 62.76 & 73.30 & 71.18 \\
Vision-OPD & 4B & 90.05 & 59.76 & 83.25 & 81.25 & 75.17 & 71.00 & 76.75 \\
OPD-V & 4B & 90.05 & 61.42 & 84.12 & 81.50 & 75.37 & 71.66 & 77.35 \\
Decomposed OPD & 4B & 89.80 & 58.70 & 82.60 & 79.50 & 73.80 & 67.82 & 75.37 \\
VAD & 4B & \underline{93.72} & 60.95 & 87.25 & 83.88 & 75.97 & 73.04 & 79.13 \\
OPD-Aha & 4B & \textbf{94.76} & \underline{62.84} & 88.12 & \underline{85.25} & \underline{77.67} & \textbf{74.58} & \underline{80.54} \\
\leaprow \textbf{LEAP} & 4B & 92.67 & \textbf{69.59} & \textbf{90.00} & \textbf{88.50} & \textbf{78.30} & \underline{74.40} & \textbf{82.24} \\
\midrule
\tabgroupavg{8}{Qwen3.5-based models (9B)}
Qwen3.5 & 9B & 86.39 & 52.78 & 85.50 & 81.75 & 71.31 & 67.50 & 74.20 \\
SFT & 9B & 90.58 & 60.71 & 80.38 & 76.75 & 71.89 & 70.44 & 75.12 \\
ZwZ2 & 9B & 90.58 & 59.53 & \underline{89.75} & 87.62 & 75.33 & \underline{73.45} & 79.38 \\
GRPO & 9B & 89.10 & 57.41 & 85.12 & 82.75 & 73.48 & 69.35 & 76.20 \\
VA-OPD & 9B & 86.91 & 62.84 & 86.12 & 82.75 & 71.17 & 70.20 & 76.67 \\
V-Zero & 9B & 89.53 & 57.04 & 85.50 & 83.88 & 75.29 & 71.52 & 77.13 \\
Vision-OPD & 9B & 91.10 & \underline{65.56} & 88.25 & 85.50 & 73.57 & 70.46 & 79.07 \\
OPD-V & 9B & 93.72 & 63.67 & 84.62 & 82.62 & 71.81 & 71.15 & 77.93 \\
Decomposed OPD & 9B & 91.40 & 60.20 & 85.80 & 83.10 & 72.70 & 69.10 & 77.05 \\
VAD & 9B & \underline{95.29} & 61.78 & 88.00 & 84.75 & 76.59 & 72.22 & 79.77 \\
OPD-Aha & 9B & \textbf{96.34} & 63.31 & 89.12 & \underline{88.12} & \textbf{78.55} & \textbf{75.21} & \underline{81.78} \\
\leaprow \textbf{LEAP} & 9B & 94.76 & \textbf{69.70} & \textbf{90.75} & \textbf{90.25} & \underline{77.38} & 72.49 & \textbf{82.56} \\
\bottomrule
\end{tabularx}
\endgroup
\end{table}

%% file: tab/table_gap.tex
\begin{table}[t]
\centering
\caption{\textbf{Mitigating the imitation gap.} Yes-bias: change from base (pp; base row: \%). Acc.\ change: tool vs.\ no tool (pp).}
\label{tab:gap}
\begingroup
\tabsetup{10}{13}{6pt}\setlength{\aboverulesep}{0.25ex}\setlength{\belowrulesep}{0.45ex}
\begin{tabularx}{\linewidth}{l *{4}{C}}
\toprule
& \multicolumn{2}{c}{Yes-bias} & \multicolumn{2}{c}{Native zoom tool} \\
\cmidrule(lr){2-3}\cmidrule(l){4-5}
Model & False claim$\downarrow$ & False alarm$\downarrow$ & Calls (\%)$\uparrow$ & Acc.\ change$\uparrow$ \\
\midrule
Base & 35.1 & 29.4 & 100.00 & +9.47 \\
\midrule
Vision-OPD & +17.4 & +14.0 & 49.74 & \textminus6.28 \\
VAD & +25.9 & +18.1 & 58.67 & \textminus2.09 \\
OPD-Aha & +14.1 & +10.1 & 22.75 & \textminus2.37 \\
OPD-V & +27.7 & +20.1 & 7.11 & \textminus7.09 \\
\leaprow \textbf{LEAP} & \textbf{+10.2} & \textbf{+7.0} & \textbf{87.26} & \textbf{+0.76} \\
\bottomrule
\end{tabularx}
\endgroup
\end{table}

%% file: sec/4_3_ablations.tex
\subsection{Ablation Studies}
\label{sec:ablations}

\input{tab/table2_components.tex}
\input{fig/fig4_gain_source.tex}
\paragraph{Contrast and acquisition reinforce each other.}
Table~\ref{tab:components} tests whether answer distillation and evidence acquisition can work alone under shared test-time acquisition.
While VPD ($\beta=4$) supplies most of the Avg$_6$ gain over the plain crop teacher ($\beta=0$, 81.88\% vs.\ 78.70\%), stronger answer-only imitation aggravates yes-bias (+12.5 false-claim points) and suppresses native zoom calls to 13.0\% (Tables~\ref{tab:bias} and~\ref{tab:native}), dropping ZoomBench from 67.46\% to 66.27\%.
Adding PEA (67.34\%) or BAT (67.57\%) alone partly recovers this drop, whereas combining both in LEAP curbs both symptoms (+10.2 points, 87.3\% calls) and lifts ZoomBench by $+3.32$ points to 69.59\% (82.24\% Avg$_6$).
Conversely, at $\beta=0$, BAT and PEA fail to help (66.86\% ZoomBench, 77.53\% Avg$_6$): acquisition training changes the view gain by $-1.63$ points at $\beta=0$ versus $+0.55$ at $\beta=4$ ($+2.18$-point interaction; Table~\ref{tab:acquisition-interaction}).

\input{tab/table3_bat.tex}
\paragraph{PEA grounds where to look.}
Table~\ref{tab:pea} tests how to supervise $q_\theta$ on ZoomBench.
Rewarding answer correctness (Outcome) drops the hit rate from 74.52\% (VPD + BAT) to 42.51\% and mean best IoU from 0.54 to 0.23, costing $2.00$ Avg$_6$ points against LEAP (Table~\ref{tab:ablation-full}).
With the full image in $H_a$, 54.44\% of correct answers with views miss the evidence ($U'$), whereas PEA's overlap reward cuts $U'$ to 18.28\%, lifts IoU to 0.68, and raises the hit rate to 79.73\% (VPD + PEA) and 78.59\% (LEAP).

\input{tab/table4_pea.tex}
\paragraph{BAT restores inspection priority.}
Localization helps only when the policy inspects where direct answers fail (Table~\ref{tab:bat}).
While the base trigger $g_\theta$ ranks its own direct-answer errors above chance (0.56 AUROC, 74.69\% recall), crop-teacher distillation (0.44), Vision-OPD-4B (0.44), and VPD (0.39 AUROC, 50.42\% recall) invert this priority below chance (Fig.~\ref{fig:gain-source}b).
Adding PEA alone (0.38 AUROC) leaves this inversion intact, whereas BAT's failure calibration restores ranking to 0.57 AUROC (VPD + BAT) and 0.62 in LEAP at 4B and 9B (83.33\% recall).

\paragraph{Tool use needs both PEA and BAT.}
Consequently, with the native zoom tool (Table~\ref{tab:native}), neither PEA (33.2\%) nor BAT (1.9\%) alone overcomes VPD's suppression (13.0\%), whereas training both in LEAP restores native calls to 87.3\% ($+65.1$-point interaction) at a 9.2\% break rate.

\takeaway{\textbf{Verifiable where-and-when supervision unlocks active distillation.} Box overlap (PEA) grounds where to look and failure calibration (BAT) restores when to look; native tool use recovers only when both are trained jointly.}

%% file: tab/table2_components.tex
\begin{table}[t]
\centering
\caption{\textbf{Answer imitation and evidence acquisition.} Qwen3.5-4B; every row, Base included, uses test-time acquisition.}
\label{tab:components}
\begingroup
\tabsetup{10}{13}{6pt}\setlength{\aboverulesep}{0.25ex}\setlength{\belowrulesep}{0.45ex}
\begin{tabularx}{\linewidth}{l *{5}{C}}
\toprule
Training & $\beta$ & BAT & PEA & ZoomBench$\uparrow$ & Avg$_6\uparrow$ \\
\midrule
Base & \tabna & \tabna & \tabna & 58.11 & 75.39 \\
Crop teacher & 0 & \tabna & \tabna & 67.46 & 78.70 \\
VPD & 4 & \tabna & \tabna & 66.27 & 81.88 \\
VPD + PEA & 4 & \tabna & $\checkmark$ & 67.34 & 81.94 \\
VPD + BAT & 4 & $\checkmark$ & \tabna & 67.57 & 81.62 \\
Crop teacher & 0 & $\checkmark$ & $\checkmark$ & 66.86 & 77.53 \\
\leaprow \textbf{LEAP} & 4 & $\checkmark$ & $\checkmark$ & \textbf{69.59} & \textbf{82.24} \\
\bottomrule
\end{tabularx}
\endgroup
\end{table}

%% file: fig/fig4_gain_source.tex
\begin{figure}[t]
\centering
\includegraphics[width=\linewidth]{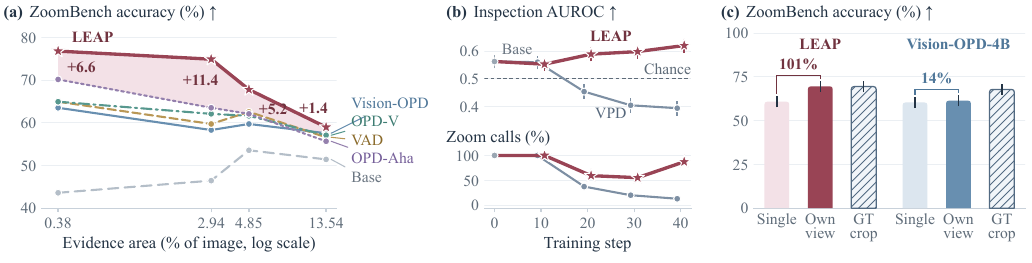}
\caption{\textbf{LEAP's lead concentrates on small evidence, and its own views match ground-truth views.} (a) On ZoomBench, LEAP's lead over the best released method concentrates on the two smallest evidence quartiles (up to $+11.4$ pp). (b) Without BAT and PEA, VPD suppresses zoom calls and drives inspection AUROC below chance during training, whereas LEAP lifts it to $0.62$. (c) With our procedure, LEAP's own views recover 101\% of its single-pass-to-GT-crop gap, versus 14\% for Vision-OPD-4B. Error bars: 95\% intervals.}
\label{fig:gain-source}
\end{figure}

%% file: tab/table3_bat.tex
\begin{table}[t]
\centering
\caption{\textbf{Restoring inspection priority.} AUROC: how well the first-token score $g_\theta$ ranks the student's own direct-answer errors. Error recall at a fixed 72.6\% inspection rate (random: 72.6).}
\label{tab:bat}
\begingroup
\tabsetup{10}{13}{6pt}\setlength{\aboverulesep}{0.25ex}\setlength{\belowrulesep}{0.45ex}
\begin{tabularx}{\linewidth}{l *{2}{C}}
\toprule
Training & AUROC$\uparrow$ & Error recall (\%)$\uparrow$ \\
\midrule
Base & 0.56 & 74.69 \\
Crop teacher & 0.44 & 51.99 \\
VPD & 0.39 & 50.42 \\
VPD + PEA & 0.38 & 52.38 \\
VPD + BAT & 0.57 & 79.69 \\
\leaprow \textbf{LEAP} & \textbf{0.62} & \textbf{83.33} \\
\bottomrule
\end{tabularx}
\endgroup
\end{table}

%% file: tab/table4_pea.tex
\begin{table}[t]
\centering
\caption{\textbf{Acquiring the relevant evidence.} Region reward on ZoomBench; all rows except VPD + PEA also train BAT. Hit: evidence center inside a view of at most 25\% of the image; $U'$: share of correct answers with views but no localization hit.}
\label{tab:pea}
\begingroup
\tabsetup{10}{13}{6pt}\setlength{\aboverulesep}{0.25ex}\setlength{\belowrulesep}{0.45ex}
\begin{tabularx}{\linewidth}{l *{3}{C}}
\toprule
PEA reward & Hit (\%)$\uparrow$ & IoU$\uparrow$ & $U'$ (\%)$\downarrow$ \\
\midrule
None (VPD + BAT) & 74.52 & 0.54 & 22.05 \\
Outcome & 42.51 & 0.23 & 54.44 \\
IoU (VPD + PEA) & \textbf{79.73} & \textbf{0.68} & 19.32 \\
\leaprow \textbf{IoU (LEAP)} & 78.59 & \textbf{0.68} & \textbf{18.28} \\
\bottomrule
\end{tabularx}
\endgroup
\end{table}

%% file: sec/4_4_evidence_acquisition.tex
\subsection{Evidence Acquisition and the Imitation Gap}
\label{sec:evidence-quality}

\paragraph{Self-acquired views match ground-truth-view accuracy.}
We next test whether LEAP's acquired views equalize access to the teacher's crop.
Across all 845 ZoomBench items, LEAP reaches 69.59\% with its self-acquisition policy (51 items answered directly) and matches its 69.47\% accuracy with the full image and ground-truth view, whereas our acquisition procedure closes only 14\% of Vision-OPD-4B's gap to its ground-truth view (Fig.~\ref{fig:gain-source}c).
Because acquired views resolve details that global encoding pools together, LEAP's lead over passive distillation concentrates on the two smallest evidence quartiles (Fig.~\ref{fig:gain-source}a); replacing those views with random crops or noise images~\citep{wang2026illusion} drops accuracy by 7.93 and 7.57 points back to the single-pass level.

\paragraph{Mitigating yes-bias and tool-call suppression.}
Equalizing access mitigates both levels of the gap (Table~\ref{tab:gap}).
At the answer level on HallusionBench~\citep{guan2024hallusionbench}, released methods increase ``yes'' answers to the false wording of paired claims (False claim) and to gold-``no'' questions (False alarm); LEAP adds the fewest on both ($+10.2$ and $+7.0$ points) and alone gains net claim verification ($+1.7$ points), though its masked abstention trails the base by 35.5 points (Table~\ref{tab:bias}).
At the behavior level, released methods call their native zoom tool on at most 58.67\% of questions and lose V$^\star$/ZoomBench accuracy when deciding to zoom (Acc.\ change), whereas LEAP calls it on 87.26\% and gains $+0.76$ points.

\takeaway{\textbf{Learned evidence acquisition mitigates the imitation gap.} Self-acquired views match ground-truth-view accuracy on ZoomBench while LEAP curbs end-to-end yes-bias and preserves 87\% native zoom calls.}

%% file: sec/4_5_remote_sensing.tex
\subsection{Transfer to Remote Sensing}
\label{sec:remote-sensing}
\input{tab/table5_remote_sensing.tex}
To further evaluate out-of-domain generalization in an extreme high-resolution, fine-grained setting, we use remote sensing as a stress testbed (Table~\ref{tab:remote-sensing}).
In the zero-shot setting, LEAP achieves competitive results, gaining $9.15$ and $7.20$ average points over Qwen3.5 at 4B and 9B and matching the domain-trained RS-OPSD-8B~\citep{jiang2026rsopsd} at 9B (52.76\% vs.\ 51.70\%).
When trained on RS-OPSD's remote-sensing data~\citep{jiang2026rsopsd}, our 4B model achieves the best overall accuracy (LEAP-RS, 53.85\%).

%% file: tab/table5_remote_sensing.tex
\begin{table}[t]
\centering
\caption{\textbf{Remote-sensing transfer.} Shared-protocol accuracy (\%).}
\label{tab:remote-sensing}
\begingroup
\tabsetup{10}{13}{6pt}\setlength{\aboverulesep}{0.25ex}\setlength{\belowrulesep}{0.45ex}
\begin{tabularx}{\linewidth}{l c *{3}{C} G}
\toprule
Model & Size & XLRS & MME-RS & LRS & \textbf{Avg} \\
\midrule
RS-OPSD & 8B & 53.40 & 61.57 & 40.12 & 51.70 \\
RS-OPD-Lite & 2B & 43.43 & 52.52 & 37.75 & 44.57 \\
\midrule
Qwen3.5 & 4B & 41.95 & 51.95 & 29.74 & 41.21 \\
Qwen3.5 & 9B & 43.80 & 61.58 & 31.30 & 45.56 \\
\leaprow \textbf{LEAP} & 4B & 51.04 & 62.79 & 37.26 & 50.36 \\
\leaprow \textbf{LEAP} & 9B & \textbf{56.22} & 65.25 & 36.80 & 52.76 \\
\leaprow \textbf{LEAP-RS} & 4B & 52.98 & \textbf{67.91} & \textbf{40.66} & \textbf{53.85} \\
\bottomrule
\end{tabularx}
\endgroup
\end{table}

%% file: sec/5_conclusion.tex
\section{Conclusion}
\label{sec:conclusion}
Passive answer imitation in privileged vision distillation introduces a two-level imitation gap: matching a crop teacher from an unresolved global view transfers confident answering behavior without perceptual access, and the distilled students show end-to-end yes-bias and suppressed native tool use.
\textbf{LEAP} shifts privileged distillation from passive answer imitation to active evidence acquisition by factorizing one policy into trigger, tool-call, and answer tokens.
It couples contrastive answer distillation (VPD) for \emph{what to answer} with verifiable rewards from the evidence region (PEA) for \emph{where to look} and from the student's direct-answer failures (BAT) for \emph{when to look}.
Across six fine-grained benchmarks, LEAP mitigates the imitation gap on both levels by cutting false affirmations below the released methods and preserving native zoom calls, while on ZoomBench its self-acquired views match the accuracy of ground-truth evidence views and its lead over passive distillation concentrates on the smallest evidence.
Our findings show that when supervision comes from privileged information, distillation should teach how to acquire evidence beyond only imitating the teacher's answering behavior.

%% file: sec/6_llm_usage.tex
\section*{\pdfbookmark[1]{LLM Usage Statement}{llm-usage}LLM Usage Statement}
In this work, we used generative AI tools (Claude Opus 5.5 and GPT 6 Astra) for language editing, improving clarity and readability, and refining the wording of the manuscript. We have not used generative AI tools for generating research data, conducting data analysis, or drawing scientific conclusions; other disclosure categories are not applicable to this work. Additionally, we used generative AI tools to suggest alternative phrasing and improve the overall presentation of the text. We have reviewed all AI-assisted work carefully. All AI-generated or AI-assisted text was checked and revised by the authors to ensure accuracy, consistency with the underlying research, and appropriate attribution of relevant sources. No AI-generated content was accepted without author review. We take responsibility for the final content of this work, including text, claims, or artifacts produced with the aid of generative AI.

%% file: sec/C_analysis.tex
\section{Analysis of Observation and Acquisition}
\label{app:leap_theory}

This section provides the formal analysis and proofs for Sec.~\ref{sec:leap}.
Appendix~\ref{app:observation-residual} decomposes imitation risk into an observation residual and a fitting error, shows how an acquired view lowers the residual, and derives what spatial overlap guarantees about the view.
Appendix~\ref{app:missing-inspection} proves that no reweighting of teacher labels can supply an inspection action the teacher never takes, which is why passive distillation leaves the trigger unsupervised and LEAP supervises acquisition from outside the teacher.
Appendix~\ref{app:leap_trigger_proof} characterizes BAT's shift-invariant objective, shows that sampled failure labels leave its population optimum unchanged, and links failure ranking to the gain of selective inspection.

\paragraph{Setup and notation.}
We analyze population risks on finite alphabets at a fixed response context, student representation, and frozen teacher.
Let $S$ denote the task state, $T_S$ the teacher's target distribution on the answer alphabet $\mathcal Y$, and $X=f(S)$ the student's effective observation: the evidence that the global encoding resolves from the clean input $x=(I,q)$, whose realization we also write $x$.
When fine-grained visual details inside $b^\star$ are lost at global resolution, distinct task states map to the same effective observation $X$ and are pooled by the student.
Under the factorized policy of Eq.~\eqref{eq:leap_policy}, the acquisition action $a=(z,u)$ is drawn from $X$ alone and is therefore conditionally independent of $S$ given $X$; at inference, sampling is replaced by a calibrated threshold and greedy decoding (Appendix~\ref{app:impl-inference}).
Finally, any observation $V$ that determines the teacher's target, such as an acquired view $H_a=(X,Z_a)$ that reproduces the teacher's crop, has zero residual and the optimal reader $\bar T_V=T_S$ (Proposition~\ref{prop:leap_projection}(d)): the target that VPD fits on the region-marked image $I^g$ is the optimum on such views, and Fig.~\ref{fig:gain-source}c tests the transfer of its answer supervision to acquired views at test time.

\subsection{Observation Residual and View Acquisition}
\label{app:observation-residual}

For any observation $V$, let $\bar T_V=\mathbb E[T_S\mid V]$ denote the pooled teacher target.
The conditional observation residual at a realization $v$ and its expectation are
\begin{equation}
\mathcal G(v)=\mathbb E[\mathrm{KL}(T_S\|\bar T_V)\mid V=v],\qquad
\mathcal G(V)=\mathbb E_{v\sim V}\,\mathcal G(v).
\label{eq:leap_conditional_residual}
\end{equation}
Stating the residual per observation $v$ is essential because $T_S$ can coincide with $\bar T_v$ at a single state while the observation pool remains ambiguous: for three equiprobable pooled states with teacher probabilities $0,\tfrac12,1$ on a binary answer, the middle state has zero pointwise discrepancy even though $\mathcal G(v)>0$; read pointwise, the equivalence in Proposition~2 of ADVISOR~\citep{weihs2021advisor} fails on this example.
A reader $p_V$ is any conditional distribution on $\mathcal Y$ given $V$, with $\mathrm{KL}$ taking values in $[0,\infty]$ and $0\log0=0$.
An acquisition action $a\sim\rho(\cdot\mid X)$ returns a view $Z_a$, augmenting the student's observation to $H_a=(X,Z_a)$, where $Z_a$ is empty for a direct answer ($z{=}0$).

\begin{leapproposition}[Projection, refinement, and acquisition]
\label{prop:leap_projection}\leavevmode
\begin{leapitems}
\item[(a)] For every reader $p_X$, Eq.~\eqref{eq:leap_projection} holds, and its minimizer is $p_X^\star=\bar T_X$, unique on observations of positive probability.
\item[(b)] $\mathcal G(x)=\operatorname{Ent}(\bar T_x)-\mathbb E[\operatorname{Ent}(T_S)\mid X=x]\le\log|\mathcal Y|$ for Shannon entropy $\operatorname{Ent}$, and $\mathcal G(x)=0$ if and only if $T_S=\bar T_x$ almost surely given $X=x$.
\item[(c)] For any augmented observation $H=(X,Z)$,
\begin{equation}
\mathcal G(x)=\mathbb E[\mathrm{KL}(T_S\|\bar T_H)\mid X=x]
+\mathbb E[\mathrm{KL}(\bar T_H\|\bar T_X)\mid X=x].
\label{eq:leap_conditional_refinement}
\end{equation}
Taking expectations over $X$ yields Eq.~\eqref{eq:leap_refinement_kl}, with strict reduction $\mathcal G(H)<\mathcal G(X)$ if and only if $\bar T_H\ne\bar T_X$ with positive probability; in particular, a view can lower the residual at $x$ only when $\mathcal G(x)>0$.
\item[(d)] If $T_S$ is almost surely a function of $H$, then $\mathcal G(H)=0$ and the unique optimal reader on $H$ is $p_H^\star=T_S$.
\item[(e)] For any acquisition policy $\rho$ and readers $p_{H_a}$,
\begin{equation}
\mathbb E\,\mathrm{KL}(T_S\|p_{H_a})
=\mathcal G(X)
-\mathbb E\sum_a\rho(a\mid X)\,\mathrm{IG}_a(X)
+\mathbb E\sum_a\rho(a\mid X)\,\mathrm{Fit}_a(X),
\label{eq:leap_acquisition}
\end{equation}
where the view's information gain and the fitting error on $H_a$ are
\[
\begin{aligned}
\mathrm{IG}_a(x)&=\mathbb E[\mathrm{KL}(\bar T_{H_a}\|\bar T_X)\mid X=x]\ge0,\\
\mathrm{Fit}_a(x)&=\mathbb E[\mathrm{KL}(\bar T_{H_a}\|p_{H_a})\mid X=x]\ge0.
\end{aligned}
\]
\end{leapitems}
\end{leapproposition}

\begin{proof}
(a) Decompose $\log(T_S/p_X)$ as $\log(T_S/\bar T_X)+\log(\bar T_X/p_X)$.
The second term depends only on $X$, so iterated expectation replaces $T_S$ with $\bar T_X$ in it:
\begin{equation}
\mathbb E\sum_y T_S(y)\log\frac{\bar T_X(y)}{p_X(y)}
=\mathbb E\sum_y\bar T_X(y)\log\frac{\bar T_X(y)}{p_X(y)}
=\mathbb E\,\mathrm{KL}(\bar T_X\|p_X).
\end{equation}
Nonnegativity and the zero condition of $\mathrm{KL}$ give the minimizer $\bar T_X$, unique on positive-probability observations.
\leapstep{b}Applying iterated expectation to $-\sum_yT_S(y)\log\bar T_x(y)$ yields the entropy difference, bounded by $\operatorname{Ent}(\bar T_x)\le\log|\mathcal Y|$ and $\operatorname{Ent}(T_S)\ge0$, with equality to zero if and only if $T_S=\bar T_x$ almost surely given $X=x$.
\leapstep{c}Apply (a) conditioned on $X=x$ to the observation $H$ and the reader that outputs $\bar T_X$.
Since both terms on the right-hand side of Eq.~\eqref{eq:leap_conditional_refinement} are nonnegative, the second term is at most $\mathcal G(x)$ and is strictly positive if and only if $\bar T_H\ne\bar T_X$ with positive probability given $X=x$.
\leapstep{d}If $T_S$ is a function of $H$, then $\bar T_H=T_S$ almost surely, so $\mathcal G(H)=0$ and (a) gives $p_H^\star=T_S$.
\leapstep{e}Fix $x$ and $a$.
Because $a$ is independent of $S$ given $X$, conditioning on $(X=x,a)$ preserves the conditional law of $S$ given $X=x$.
Applying (c) with $H=H_a$ gives
\[
\mathbb E[\mathrm{KL}(T_S\|\bar T_{H_a})\mid X=x,a]=\mathcal G(x)-\mathrm{IG}_a(x),
\]
and applying (a) on $H_a$ adds $\mathrm{Fit}_a(x)$.
Taking expectations over $a\sim\rho(\cdot\mid X)$ and $X$ yields Eq.~\eqref{eq:leap_acquisition}.
\end{proof}

\paragraph{How the decomposition governs LEAP.}
Eq.~\eqref{eq:leap_acquisition} shows how each factor of LEAP's policy in Eq.~\eqref{eq:leap_policy} enters the imitation risk.
The answer policy $\pi_\theta(y\mid H_a)$ acts solely on the fitting error $\mathrm{Fit}_a(X)$, the term that VPD targets, while the tool-call policy $q_\theta(u\mid x,1)$ selects the region $b$ that determines $\mathrm{IG}_a(X)$, for which PEA's overlap reward is a spatial proxy.
The trigger $g_\theta(z{=}1\mid x)$ then weights the net per-input risk reduction over a direct answer with fitting error $\mathrm{Fit}_0(x)$,
\[
\mathrm{IG}_a(x)-\bigl(\mathrm{Fit}_a(x)-\mathrm{Fit}_0(x)\bigr),
\]
whose information gain $\mathrm{IG}_a(x)$ is positive only when $\mathcal G(x)>0$, for which BAT's direct-answer failures are an observable proxy.
A view thus lowers the imitation risk only if its information gain exceeds the increase in fitting error, consistent with passive students losing accuracy when they zoom (Fig.~\ref{fig:teaser}c, Table~\ref{tab:gap}).

\paragraph{Pooling, commitment, and yes-bias.}
Proposition~\ref{prop:leap_projection}(a) shows that behavior invariant across an observation pool transfers intact, whereas evidence-dependent targets are replaced by their conditional average $\bar T_x$.
This pooling underlies both symptoms of the answer-level imitation gap when a student imitates a crop teacher from the global view.
First, when the teacher, which resolves the evidence in $I^\star$, gives abstention little mass, the optimal student on $X=x$ inherits the pool's average of it, so it commits even when the evidence is unresolved, consistent with the abstention it loses on masked evidence (Table~\ref{tab:bias}).
Second, consider any pair of a true claim $S^+$ and a false claim $S^-$ whose distinguishing visual detail inside $b^\star$ is unresolved at global resolution, so that $f(S^+)=f(S^-)=x$.
Because any single-pass student reader $p_X$ depends only on $X$, it must assign the same affirmative probability to both claims:
\[
\Pr_{p_X}(\text{Yes}\mid S^-)=p_x(\text{Yes})=\Pr_{p_X}(\text{Yes}\mid S^+).
\]
Consequently, whatever raises $p_x(\text{Yes})$ on unresolved pools raises false affirmations together with true ones and, under a forced yes/no answer, cannot improve claim verification on the pair, consistent with the similar increases on true and false claims in Table~\ref{tab:bias}.
Acquiring a crop $Z_a$ that resolves $b^\star$ separates $H_a(S^+)$ from $H_a(S^-)$, yielding $\mathrm{IG}_a(x)>0$ whenever the teacher distinguishes the two claims.

\paragraph{Perceptual access versus missing information.}
Whenever $b^\star$ is determined by the clean input $(I,q)$, the teacher's crop $I^\star=\operatorname{Crop}(I,b^\star)$ and target $T_S$ are functions of $(I,q)$ at the given response context.
By Proposition~\ref{prop:leap_projection}(d), the residual on the raw input $(I,q)$ is zero, so $\mathcal G(X)>0$ arises entirely from the lossy global encoding $X=f(S)$.
This formalizes the distinction between privileged vision distillation, where the student can recover $I^\star$ from its own image via a zoom call, and distillation under privileged text, where the privileged context is absent from the student's input.
A text prefix that the student generates from $X$ alone cannot reduce $\mathcal G(X)$ either: since the prefix is drawn conditionally on $X$, $S$ remains conditionally independent of it given $X$, leaving the observation pools unchanged.

\paragraph{Geometric guarantees of spatial overlap.}
When a candidate box $b$ matches the teacher's region ($\operatorname{IoU}(b,b^\star)=1$), the returned view reproduces $I^\star$ and $\mathcal G(H)=0$ by Proposition~\ref{prop:leap_projection}(d).
More generally, since $|b\cap b^\star|\ge\kappa|b\cup b^\star|$ and $\max(|b|,|b^\star|)\le|b\cup b^\star|\le|b\cap b^\star|/\kappa$, any box with $\operatorname{IoU}(b,b^\star)\ge\kappa>0$ satisfies
\[
\frac{|b\cap b^\star|}{|b^\star|}\ge\kappa,\qquad
\frac{|b\cap b^\star|}{|b|}\ge\kappa,\qquad
|b|\le\frac{|b^\star|}{\kappa}.
\]
The first inequality guarantees that $b$ covers at least a fraction $\kappa$ of the privileged region $b^\star$; the second guarantees that the covered part fills at least a fraction $\kappa$ of the cropped view $b$, which limits the resolution lost to an oversized box.
Furthermore, for axis-aligned boxes, $\operatorname{IoU}(b,b^\star)>\tfrac12$ guarantees that $b$ contains the center of $b^\star$ (the center condition of the Hit criterion in Table~\ref{tab:pea}), because any box missing the center of $b^\star$ can intersect at most half of its area.
IoU is therefore a verifiable spatial proxy for the residual reduction that needs no teacher rollouts on candidate crops: a higher overlap raises both area bounds, and the gain from acquired views grows with it (Table~\ref{tab:grounding-details}).

\paragraph{Extension to Jensen--Shannon divergence.}
Proposition~\ref{prop:leap_projection}(a,e) uses forward KL's chain rule, but two properties that motivate view acquisition carry over to the Jensen--Shannon divergence $\mathrm{JS}_{1/2}$ minimized in Eq.~\eqref{eq:leap_vpd}.
First, the minimum risk $\mathcal R_{\mathrm{JS}}(V)=\inf_{p_V}\mathbb E\,\mathrm{JS}_{1/2}(T_S,p_V)$ of any observation $V$ is monotonically non-increasing under view acquisition:
\[
\mathcal R_{\mathrm{JS}}(X,Z)\;\le\;\mathcal R_{\mathrm{JS}}(X),
\]
because every reader of $X$ is a special case of a reader of $(X,Z)$ that ignores $Z$.
Second, because $\mathrm{JS}_{1/2}(P,Q)=0$ if and only if $P=Q$, the minimum risk $\mathcal R_{\mathrm{JS}}(V)$ vanishes if and only if $T_S=\bar T_V$ almost surely, in which case the unique optimal reader on a resolved view is $p_V^\star=T_S$, identical to forward KL.
On an unresolved pool $X=x$ of point-mass targets with unequal frequencies $r(y)={\Pr(T_S=\delta_y\mid X=x)}$, the $\mathrm{JS}_{1/2}$ optimum is a pooled distribution sharper than the mean: writing $u(t)=t\log t-(1+t)\log(1+t)$ gives
\[
\mathbb E[\mathrm{JS}_{1/2}(T_S,q)\mid X=x]=\log2+\tfrac12\sum_yr(y)\,u(q(y)),
\]
which is convex in $q$ and whose unique minimizer satisfies $r(y)\log(1+1/q^\star(y))=\lambda$ on the support of $r$, or
\begin{equation}
q^\star(y)=\frac{1}{\exp(\lambda/r(y))-1},\qquad \sum_yq^\star(y)=1.
\label{eq:leap_js_optimum}
\end{equation}
Because $q^\star(y)/r(y)$ is proportional to $w/(e^w-1)$ with $w=\lambda/r(y)$, it strictly increases with $r(y)$, overweighting the pool's dominant answer (for binary answers, $r=0.7$ yields $q^\star=0.817$ and $r=0.9$ yields $q^\star=0.998$).
Trained at $\beta=0$, forward KL, $\mathrm{JS}_{1/2}$, and reverse KL perform alike and all raise false affirmations over the base model (Tables~\ref{tab:ablation-full} and~\ref{tab:ablation-bias}).

\subsection{Missing Inspection Actions}
\label{app:missing-inspection}

A teacher that already holds the cropped evidence never needs to invoke an inspection action that a student without the crop requires.
This is the missing-action obstruction of ELF~\citep{walsman2023impossibly}, which we state directly for the distillation objectives in Sec.~\ref{sec:leap}.

\begin{leapproposition}[No reweighting supplies a missing action]
\label{prop:leap_missing}
Let $T_S(y_0)=0$ almost surely given $X=x$.
\begin{leapitems}
\item[(a)] For any fixed weights $W\ge0$ with a positive mean $\mathbb E[W\mid X=x]$, every minimizer of the $W$-weighted forward KL or $\mathrm{JS}_{1/2}$ at $x$ assigns zero probability to $y_0$, and any reader with positive probability on $y_0$ incurs infinite $W$-weighted reverse KL.
\item[(b)] If a context has probability zero under the training rollout distribution, the distillation objective contains no term for the policy in that context.
\item[(c)] Let a uniform binary state $Y\in\{0,1\}$ be independent of $X$, with actions answer $0$, answer $1$, and inspect, where inspection costs $0<c<\tfrac12$ and reveals $Y$ before answering, and a correct answer earns $1$.
A teacher that observes $Y$ answers directly with return $1$, so its action distribution assigns zero mass to inspect under every state distribution.
Every student that reads only $X$ earns at most $1-c$, only a student that inspects with probability $1$ attains $1-c$, and any student that never inspects earns at most $\tfrac12$.
\end{leapitems}
\end{leapproposition}

\begin{proof}
(a) For forward KL, the $W$-weighted objective at $x$ is equivalent to Proposition~\ref{prop:leap_projection}(a) under the conditional law tilted by $W$, whose unique minimizer is the tilted average
\[
\mathbb E[WT_S\mid X=x]/\mathbb E[W\mid X=x],
\]
which assigns zero mass to $y_0$.
For reverse KL, $p(y_0)>0$ makes $\mathrm{KL}(p\|T_S)=\infty$ on a set of positive weight.
The partial derivative of $F(q)=\mathbb E[W\,\mathrm{JS}_{1/2}(T_S,q)\mid X=x]$ is
\[
\partial F/\partial q(y)=\tfrac12\mathbb E[W\log(2q(y)/(T_S(y)+q(y)))\mid X=x].
\]
Choose $y_1$ with ${\mathbb E[WT_S(y_1)\mid X=x]>0}$.
If ${q(y_0)>0}$,
\[
\partial F/\partial q(y_1)-\partial F/\partial q(y_0)
=\tfrac12\mathbb E[W\log(q(y_1)/(T_S(y_1)+q(y_1)))\mid X=x]<0
\]
(possibly $-\infty$), so shifting mass from $y_0$ to $y_1$ strictly reduces $F$, and $q$ cannot be a minimizer.
\leapstep{b}A context with zero rollout weight contributes zero to the expected loss.
\leapstep{c}A student that inspects with probability $\alpha$ and then answers correctly with probability $s\le1$ achieves expected return $(1-\alpha)/2+\alpha(s-c)\le1-c$ because $1-c>\tfrac12$, with equality if and only if $\alpha=s=1$.
\end{proof}

\paragraph{Suppression of native tool calls and abstention.}
Proposition~\ref{prop:leap_missing} explains both behaviors that passive distillation suppresses: native tool invocation (Table~\ref{tab:native}) and abstention (Table~\ref{tab:bias}).
Whether a passive distillation objective visits the tool-declared context or omits it, it never rewards inspection: if visited, a teacher that answers directly from the crop assigns zero mass to \texttt{<tool\_call>}, so item~(a) drives its probability to zero; if omitted, as in our VPD rows (Appendix~\ref{app:prompts}), item~(b) leaves the trigger unsupervised while training on direct-answer rollouts shifts the shared weights toward answering directly.
Item~(a) likewise governs the answer tokens: any answer to which the teacher assigns zero mass receives zero probability from every minimizer of a weighted forward KL or $\mathrm{JS}_{1/2}$ and incurs an infinite reverse-KL penalty; under forward KL, a teacher mass of at most $\varepsilon$ on any token bounds the optimal student's mass on it by $\varepsilon$, so a teacher that rarely abstains on resolved crops yields a student that commits on unresolved inputs.

\paragraph{Why adaptive reweighting cannot recover inspection.}
Item~(a) holds for any fixed nonnegative weighting $W$ of teacher labels, including importance weights and ADVISOR's adaptive imitation weights~\citep{weihs2021advisor}.
In ELF's delayed-information example~\citep{walsman2023impossibly}, an auxiliary policy can match the expert exactly at the entrance, setting ADVISOR's imitation weight to $1$ and its RL coefficient to $0$ there so that the loss minimizer retains the low-return action.
Reweighting teacher labels therefore cannot identify the missing action: learning it requires supervision from outside the teacher on clean inputs, which PEA and BAT supply from the privileged box $b^\star$ and the student's online direct-answer failures; in item~(c), active inspection lifts the student's return from the uninspected ceiling $\tfrac12$ to the achievable optimum $1-c$.

\subsection{The Centered Trigger Update and Inspection Priority}
\label{app:leap_trigger_proof}

Given failure fractions $e_i\in[0,1]$ whose batch mean $\bar e=N^{-1}\sum_ie_i$ lies in $(0,1)$ and finite trigger margins $m_i=\operatorname{logit}g_\theta(x_i)$, binary cross-entropy with soft target $e_i$ and offset-adjusted logit $m_i+\delta$ takes the form
\[
\log(1+\exp(m_i+\delta))-e_i(m_i+\delta).
\]
BAT's offset-adjusted objective and its profile over the common offset $\delta$ are
\begin{align}
F(m,\delta)&=\frac1N\sum_i\bigl[\log(1+\exp(m_i+\delta))-e_i(m_i+\delta)\bigr],\notag\\
L_{\mathrm{prof}}(m)&=\min_\delta F(m,\delta).
\label{eq:leap_profile}
\end{align}

\begin{leapproposition}[Properties of the BAT objective]
\label{prop:leap_bat}\leavevmode
\begin{leapitems}
\item[(a)] The minimizing offset $\delta^\star(m)$ is the unique solution of the calibration equation
\begin{equation}
\frac1N\sum_i\sigma(m_i+\delta)=\bar e.
\label{eq:leap_bias}
\end{equation}
\item[(b)] $L_{\mathrm{prof}}$ is convex, invariant to common margin shifts, $L_{\mathrm{prof}}(m+t\mathbf 1)=L_{\mathrm{prof}}(m)$, and has the gradient $\partial L_{\mathrm{prof}}/\partial m_i=(\sigma(m_i+\delta^\star)-e_i)/N$.
When actor and rollout margins coincide and $\delta=\delta^\star$, $c_i=d_i$ and $\mathcal L_{\mathrm{BAT}}$ in Eq.~\eqref{eq:leap_bat_loss} has the exact gradient of $L_{\mathrm{prof}}(m^\theta)$.
\item[(c)] If the trigger reads only $X$, the minimizers of the population objective $\inf_\delta\mathbb E\,F(m,\delta)$, the large-batch limit of $L_{\mathrm{prof}}$, over all margin functions $m(x)$ satisfy, for the failure probability $e_S$ of a sampled direct answer,
\[
\sigma(m(x)+\delta)=e(x),\quad e(x)=\mathbb E[e_S\mid X=x],
\]
wherever $0<e(x)<1$ (the adjusted logit diverges where $e(x)\in\{0,1\}$), so that $m^\star(x)=\operatorname{logit}e(x)+\kappa$ up to a free common shift $\kappa$.
Moreover, because $F(m,\delta)$ is affine in the targets, replacing $e_S$ with the $K$-sample failure fraction of Eq.~\eqref{eq:leap_bat_target}, whose conditional mean is $e_S$, leaves this objective and its minimizers $m^\star(x)$ unchanged for every $K\ge1$. For batches of $N\ge2$ independent inputs, the minimizers of $\mathbb E\,L_{\mathrm{prof}}$ follow the stochastic order of the targets, which is the order of $e(x)$ for $K=1$.
\item[(d)] Any strictly increasing function of $e(x)$, in particular $\sigma(m^\star(x))$, maximizes the population AUROC of ranking direct-answer failures above non-failures among all inspection scores that read only $X$.
\end{leapitems}
\end{leapproposition}

\begin{proof}
(a) $\partial F/\partial\delta=N^{-1}\sum_i(\sigma(m_i+\delta)-e_i)$ is continuous and strictly increasing in $\delta$ from $-\bar e<0$ to $1-\bar e>0$, so its unique root $\delta^\star$ minimizes the strictly convex $F(m,\cdot)$.
\leapstep{b}$F$ is jointly convex, and partial minimization over $\delta$ preserves convexity.
Since $\delta^\star(m+t\mathbf 1)=\delta^\star(m)-t$, $L_{\mathrm{prof}}(m+t\mathbf 1)=L_{\mathrm{prof}}(m)$.
By the envelope theorem ($\partial F/\partial\delta=0$ at $\delta^\star$), $\partial L_{\mathrm{prof}}/\partial m_i=\partial F/\partial m_i\big|_{\delta=\delta^\star}=(\sigma(m_i+\delta^\star)-e_i)/N$.
By Eq.~\eqref{eq:leap_bias}, $\sum_id_i=0$ at $\delta^\star$, so $c_i=d_i$ and differentiating Eq.~\eqref{eq:leap_bat_loss} at $m^\theta=\bar m$ gives $\nabla_\theta\mathcal L_{\mathrm{BAT}}=\nabla_\theta L_{\mathrm{prof}}(m^\theta)$.
\leapstep{c}Binary cross-entropy is the Bernoulli cross-entropy, so Proposition~\ref{prop:leap_projection}(a) applied to $\operatorname{Ber}(e_S)$ yields the unique optimal reader $\operatorname{Ber}(e(x))$, which corresponds to $m^\star(x)=\operatorname{logit}e(x)+\kappa$.
Since the $K$-sample fraction $e_i$ has conditional mean $e_{S_i}$ for any $K\ge1$ and $F(m,\delta)$ is affine in the targets, iterated expectation leaves $\mathbb E\,F(m,\delta)$ unchanged at every $(m,\delta)$, hence also its infimum over $\delta$ and its minimizers. With the rest of a batch fixed, Eq.~\eqref{eq:leap_bias} makes the residual $\sigma(m_i+\delta^\star)-e_i$ increase with $m_i$ and decrease with $e_i$; at a minimizer its expectation vanishes at every $x$, so stochastically larger targets need larger margins.
\leapstep{d}For any pair of observations $(x,x')$, ranking $x$ above $x'$ contributes $e(x)(1-e(x'))$ to the pairwise AUROC numerator, ranking $x'$ above $x$ contributes $e(x')(1-e(x))$, and a tie contributes their mean; the denominator does not depend on the score.
Since $e(x)(1-e(x'))\ge e(x')(1-e(x))$ if and only if $e(x)\ge e(x')$, any score strictly increasing in $e(x)$ maximizes the contribution of every pair.
\end{proof}

\paragraph{Decoupling relative priority from the inspection budget.}
Proposition~\ref{prop:leap_bat}(b--d) shows why BAT centers its residual update: both the failure-ranking AUROC and any fixed-share routing rule depend only on relative margin differences $m(x)-m(x')$.
Standard uncentered cross-entropy couples this ranking gradient with a common shift that pulls the mean trigger probability toward the batch failure rate $\bar e$.
By profiling out $\delta^\star$, $\mathcal L_{\mathrm{BAT}}$ keeps its margin gradient in the zero-sum subspace $\mathbf 1^\top c=0$, supervising relative inspection priority while leaving the overall inspection rate to the test-time threshold $\tau$ (Appendix~\ref{app:impl-inference}).

\paragraph{From failure ranking to selective inspection gain.}
Fix a checkpoint and a paired evaluation protocol, and let $e_0(x)$ and $e_1(x)$ denote the direct-answer and post-inspection failure probabilities at input $x$, with conditional correction rate $f(x)=\Pr(\text{fixed}\mid\text{direct wrong},x)$ and corruption rate $b(x)=\Pr(\text{broken}\mid\text{direct right},x)$.

\begin{leapcorollary}[Inspection gain and routing regret]
\label{cor:leap_gain}\leavevmode
\begin{leapitems}
\item[(a)] The expected accuracy gain from inspecting input $x$ is $\Delta(x)=e_0(x)-e_1(x)$, which equals $e_0(x)f(x)-{(1-e_0(x))}b(x)\le e_0(x)$.
\item[(b)] Suppose that $f$ and $b$ are common across inputs with ${\kappa=f+b>0}$, so $\Delta(x)=\kappa e_0(x)-b$.
For any routing rule $w(x)\in[0,1]$ that reads only $x$, with overall inspection rate $\alpha=\mathbb E[w(X)]$, let
\[
\begin{aligned}
r_{\mathrm{err}}&=\Pr(\text{inspect}\mid\text{direct wrong}),\\
r_{\mathrm{ok}}&=\Pr(\text{inspect}\mid\text{direct right}).
\end{aligned}
\]
Compared with random routing at the same budget $\alpha$, whose gain is $\Delta_{\mathrm{rand}}(\alpha)=\alpha\,\mathbb E[\Delta(X)]$, the selective routing gain satisfies
\begin{equation}
\mathbb E[w(X)\Delta(X)]-\Delta_{\mathrm{rand}}(\alpha)=\kappa\,\operatorname{Cov}\bigl(w(X),e_0(X)\bigr)
=\kappa\,\bar e_0(1-\bar e_0)\bigl(r_{\mathrm{err}}-r_{\mathrm{ok}}\bigr),
\end{equation}
where $\bar e_0=\mathbb E[e_0(X)]\in(0,1)$.
Thus selective routing outperforms random selection at the same budget if and only if $r_{\mathrm{err}}>r_{\mathrm{ok}}$, and inspecting the inputs with the highest $e_0(x)$ maximizes total gain at any fixed inspection share.
\item[(c)] For $N$ inputs of which $k$ are inspected, suppose that ${|\Delta_i-(\kappa e_{0i}-b)|\le\zeta}$ with $\kappa\ge0$, let ${s_i=\sigma(m_i+\delta)}$ for any common $\delta$, and let $\hat A$ and $A^\star$ be the top-$k$ sets ranked by $s$ and by $\Delta$.
Then the per-input routing regret satisfies
\begin{equation}
\frac1N\Bigl(\sum_{i\in A^\star}\Delta_i-\sum_{i\in\hat A}\Delta_i\Bigr)
\le\kappa\sqrt{\frac{\mathcal E}{2}}+\frac{2\min(k,N-k)}{N}\zeta,
\label{eq:leap_regret}
\end{equation}
where $\mathcal E=N^{-1}\sum_i\mathrm{KL}(\operatorname{Ber}(e_{0i})\|\operatorname{Ber}(s_i))$.
\end{leapitems}
\end{leapcorollary}

\begin{proof}
(a) By total probability, $e_1(x)=e_0(x)(1-f(x))+(1-e_0(x))b(x)$, which rearranges to $\Delta(x)=e_0(x)f(x)-(1-e_0(x))b(x)\le e_0(x)$.
\leapstep{b}Under common $f$ and $b$, $\Delta(X)=\kappa e_0(X)-b$, so $\mathbb E[w(X)\Delta(X)]-\alpha\,\mathbb E[\Delta(X)]=\kappa\,\operatorname{Cov}(w(X),e_0(X))$.
Since
\[
\begin{aligned}
\bar e_0 r_{\mathrm{err}}&=\mathbb E[w(X)e_0(X)],\\
(1-\bar e_0)r_{\mathrm{ok}}&=\mathbb E[w(X)(1-e_0(X))],
\end{aligned}
\]
the covariance is $\bar e_0(1-\bar e_0)(r_{\mathrm{err}}-r_{\mathrm{ok}})$, and ranking by $e_0(x)$ maximizes $\mathbb E[w(X)e_0(X)]$ subject to $\mathbb E[w(X)]=\alpha$.
\leapstep{c}Terms on $A^\star\cap\hat A$ and the constant offset $b$ cancel between the two sums of size $k$.
Since $\sum_{i\in\hat A}s_i\ge\sum_{i\in A^\star}s_i$, the regret is bounded by
\[
\kappa N^{-1}\sum_i|e_{0i}-s_i|+N^{-1}|A^\star\triangle\hat A|\zeta.
\]
Pinsker's inequality bounds each $|e_{0i}-s_i|$ in the sum by $\sqrt{\tfrac12\mathrm{KL}(\operatorname{Ber}(e_{0i})\|\operatorname{Ber}(s_i))}$; Jensen's inequality and $|A^\star\triangle\hat A|\le2\min(k,N-k)$ then yield Eq.~\eqref{eq:leap_regret}.
\end{proof}

Corollary~\ref{cor:leap_gain} connects BAT's failure-ranking objective directly to the inspection-priority results in Table~\ref{tab:bat} and Fig.~\ref{fig:when-to-look}.
By item~(a), the direct-answer failure rate bounds the expected gain from above, and inputs that the student already answers correctly ($e_0(x)=0$) have non-positive gain $\Delta(x)=-b(x)\le0$.
Under item~(b)'s common $f$ and $b$, the gain increases linearly with $e_0(x)$, and a trigger outperforms random selection at the same budget if and only if $r_{\mathrm{err}}>r_{\mathrm{ok}}$, that is, when its error recall exceeds its overall inspection rate (Table~\ref{tab:bat}).
In Fig.~\ref{fig:when-to-look}b, VPD without BAT yields $r_{\mathrm{err}}<r_{\mathrm{ok}}$ ($-6.1$ and $-4.7$ points for VPD and VPD + PEA), whereas adding BAT flips $r_{\mathrm{err}}-r_{\mathrm{ok}}$ positive ($+7.0$ points for VPD + BAT and $+17.9$ points for LEAP); inspecting in the order of LEAP's score captures $77\%$ of the inspection gain using half of the questions, against $50\%$ for random selection (Fig.~\ref{fig:when-to-look}c).
Finally, item~(c) bounds the top-$k$ routing regret at any inspection budget by BAT's excess cross-entropy $\mathcal E$ and the heterogeneity $\zeta$ of recoverability.

%% file: sec/A_additional_results.tex
\section{Additional Results}
\label{app:additional-results}

This section details the diagnostic, ablation, and inference-cost analyses supporting Secs.~\ref{sec:leap_gap} and~\ref{sec:main-results}--\ref{sec:evidence-quality}.

\subsection{Imitation-Gap Diagnostics}
\label{app:diagnostics}
\input{tab/supp1_bias.tex}
\input{tab/supp2_native.tex}

Tables~\ref{tab:bias} and~\ref{tab:native} detail the two levels of the imitation gap summarized in Table~\ref{tab:gap} and Secs.~\ref{sec:leap_gap} and~\ref{sec:evidence-quality}: end-to-end yes-bias and behavioral tool-call suppression.
Their same-pipeline rows evaluate the Table~\ref{tab:components} checkpoints, and the training curves in Fig.~\ref{fig:gain-source}b follow the same audit protocol.

\paragraph{End-to-end yes-bias.}
Table~\ref{tab:bias} audits final answers on HallusionBench~\citep{guan2024hallusionbench} and 200 evidence-masked ZoomBench items: Says Yes is the affirmative rate across 815 question--image pairs, each phrased once as a true and once as a false claim; HB false alarm is the affirmative rate on 642 gold-No questions; net claim verification averages accuracy over the true and false wordings; and masked abstention is the abstention rate on the evidence-masked items.
Across all four released 4B students, passive distillation raises false affirmations by $+14.1$ to $+27.7$ points and false alarms by $+10.1$ to $+20.1$ points without improving net claim verification ($-0.1$ to $-2.3$ points), whereas LEAP cuts both false-affirmation metrics below every released student and raises net claim verification by $+1.7$ points over the base model.
The released students' false alarms also rise on POPE~\citep{li2023pope} at 4B and 9B against their same-size Qwen3.5 bases (Sec.~\ref{sec:leap_gap}; 9,000 questions from the random, popular, and adversarial COCO~\citep{lin2014coco} splits, scored by rule with V-Zero's evaluation code~\citep{sun2026vzero} and thinking disabled).
Even in a no-image control where each of the 80 distinct POPE questions is asked five times without any image, the released students answer Yes in 32.2\% to 93.2\% of cases (vs.\ 2.8\% and 0.0\% for the 4B and 9B base models), whereas LEAP-4B cuts this unconditional Yes rate to 12.8\%.

\paragraph{Behavioral tool-call suppression.}
Table~\ref{tab:native} audits native tool use across the four perception benchmarks (with Forced hit and Model decides on V$^\star$ and ZoomBench): Zoom calls count valid native calls over all questions and over tool-free errors; Forced hit is the share of first forced calls whose box contains the evidence center; Fix and Break are the shares of tool-free errors corrected and tool-free correct answers broken when calling; Recovery is the paired difference from the base model in tool-assisted error recovery; and Model decides is the accuracy change over tool-free answering when the model decides whether to zoom.
While forced localization still finds the evidence ($74.02\%$--$76.27\%$ hit rate), the four released 4B students invoke their native zoom tool on only 9\% to 67\% of their tool-free errors (vs.\ 88\% for LEAP and 100\% for the base model), break 12.6\% to 32.6\% of their correct answers when zooming, and lose $2.1$ to $7.1$ points when deciding autonomously.
By contrast, LEAP calls its native zoom tool on $87.26\%$ of questions, achieves the lowest break rate ($9.2\%$), and gains $+0.76$ points from autonomous zooming.
Extending this audit to RS-OPSD-8B~\citep{jiang2026rsopsd}, a Qwen3-VL-8B~\citep{bai2025qwen3vl} student of privileged self-distillation on remote sensing, shows the same suppression: it calls its native zoom tool on 36.8\% of V$^\star$ and ZoomBench questions against 82.7\% for its base.

\subsection{Complete Ablations}
\label{app:complete-ablations}
\input{tab/supp_ablation_full.tex}
\input{tab/supp_target_choice.tex}
\input{tab/supp_acquisition_interaction.tex}
\input{tab/supp_ablation_bias.tex}

Table~\ref{tab:ablation-full} reports the six-benchmark breakdown of all ablation variants, and Tables~\ref{tab:hyperparameters}--\ref{tab:ablation-bias} analyze target contrast, its interaction with acquisition training, and answer bias across distillation settings (paired intervals in Appendix~\ref{app:intervals}).
In Table~\ref{tab:ablation-full}, ``Crop teacher'' denotes a student distilled on the plain crop target ($\beta=0$) rather than the frozen teacher itself, and ``LEAP, $\beta=0$'' corresponds to the ``Crop teacher + BAT + PEA'' row of the other tables.
Across Tables~\ref{tab:ablation-full} and~\ref{tab:acquisition-interaction}, contrastive distillation and evidence acquisition reinforce each other: without target contrast ($\beta=0$), training PEA and BAT does not improve view gains ($-0.06$ vs.\ $+1.57$ without them), whereas pairing them with contrastive VPD ($\beta=4$) lifts the view gain from $+1.28$ to $+1.83$ points ($+2.18$-point positive interaction).
Increasing null-view contrast (a crop-sized image filled with the crop's rounded mean color~\citep{qiu2026opdaha}) from $\beta=0$ to $2$ to $4$ monotonically raises LEAP from $77.53\%$ to $81.38\%$ and $82.24\%$ Avg$_6$, whereas replacing the null view with VAD's degraded crop~\citep{zhang2026vad} at $\beta=4$ drops Avg$_6$ by $1.35$ points and raises false affirmations (Tables~\ref{tab:hyperparameters} and~\ref{tab:ablation-bias}).
Across divergences at $\beta=0$, forward KL ($77.41\%$), JSD ($77.53\%$), and reverse KL ($77.67\%$) perform comparably (Table~\ref{tab:ablation-full}), whereas at $\beta=4$ JSD remains stable while KL runs exceed the gradient threshold.

\input{fig/figS_when.tex}
\subsection{Inspection Priority}
\label{app:inspection-priority}
\input{tab/supp_inspection_priority.tex}

Tables~\ref{tab:bat} and~\ref{tab:behaviour-details} evaluate how well the first-token inspection score $g_\theta(x)$ prioritizes questions that need zooming across a 2,546-question audit set.
The inspection-decision AUROC treats the policy's own direct-answer errors as positives (computed within each benchmark and pooled over positive--negative pairs), while AUROC$_{\mathrm{fix}}$ treats zoom-routed questions with wrong-to-right corrections as positives.
Under passive distillation both rankings fall below the base model's and below chance (VPD: $0.39$ AUROC and $0.32$ AUROC$_{\mathrm{fix}}$; VPD + PEA: $0.38$ and $0.35$), whereas BAT lifts both above chance, to $0.62$ AUROC and $0.60$ AUROC$_{\mathrm{fix}}$ in LEAP (Table~\ref{tab:behaviour-details}).
At a matched $72.6\%$ inspection rate (VPD's overall rate), thresholding $g_\theta(x)$ within each benchmark yields error recalls of 50.4\%, 50.9\%, 80.1\%, and 84.2\% for VPD, VPD + PEA, VPD + BAT, and LEAP, preserving the Table~\ref{tab:bat} ordering.

\paragraph{Controlled mixtures at a fixed threshold.}
To verify that BAT separates questions that need inspection from those that do not at the calibrated threshold $\tau$, we collect both kinds of questions post hoc and mix them in a controlled experiment (Fig.~\ref{fig:when-to-look}).
A question needs inspection when the model's direct answer is wrong; with $r_{\mathrm{err}}$ and $r_{\mathrm{ok}}$ the rates at which the two groups are inspected, we vary their proportions while keeping each model's calibrated threshold $\tau$ fixed.
Because the trigger scores each question separately, the inspection rate of a mixture is the proportion-weighted mean of $r_{\mathrm{err}}$ and $r_{\mathrm{ok}}$, and a trigger that ignores need, such as random selection, inspects both groups at the same rate ($r_{\mathrm{err}}=r_{\mathrm{ok}}$, the dashed diagonal in Fig.~\ref{fig:when-to-look}a).
Fig.~\ref{fig:when-to-look}a traces both rates across thresholds and marks each model's calibrated threshold $\tau$: at $\tau$, LEAP-4B inspects on average 90.3\% of the questions that need inspection versus 72.5\% of the rest, exceeding the $72.5\%$ random level by $+17.9$ $[9.2,25.4]$ points (95\% interval, resampled as in Appendix~\ref{app:intervals}).
With $\tau$ fixed, its inspection rate thus falls from 90.3\% to 72.5\% as the share of questions that need no inspection grows from 0 to 100\%, and its curve stays above the random diagonal at every other threshold.
Fig.~\ref{fig:when-to-look}b compares the gap $r_{\mathrm{err}}-r_{\mathrm{ok}}$ at $\tau$ across the trainings of Table~\ref{tab:bat}, where zero denotes random selection.
Its sign, which Corollary~\ref{cor:leap_gain}(b) ties to the gain over random selection, follows BAT among the VPD trainings: it is negative without BAT ($-6.1$ and $-4.7$ points for VPD and VPD + PEA) and positive with it ($+7.0$ and $+17.9$ points for VPD + BAT and LEAP).
In Fig.~\ref{fig:when-to-look}c, questions are inspected in order of LEAP's inspection score while uninspected questions below $\tau$ (star) keep their direct answers: inspection lifts accuracy from 76.6\% to 80.9\%, and inspecting only the top-scoring half of questions already captures $77\%$ $[61.6,96.8]$ of this gain, versus 50\% for a random half.

\subsection{Region Selection}
\label{app:region-selection}
\input{tab/supp_behaviour_details.tex}

Table~\ref{tab:grounding-details} extends Table~\ref{tab:pea} with view-compactness and ground-truth-crop recovery metrics on ZoomBench.
Across both tables, localization is measured over ZoomBench items with acquired views (share: Views): Area is the median image share of the largest view, and a Hit requires an acquired view to contain the evidence center and the largest view to cover at most 25\% of the image.
Among correctly answered items with views, $U'$ measures the ungrounded share without a hit (denominators: 526, 507, 528, and 547 across the four rows); Gain is the paired Avg$_6$ improvement over the same checkpoint's single pass, while Closed and GT gain report the share of the single-pass-to-GT-crop gap recovered by self-acquired views and the remaining difference to GT crops.
Rewarding final-answer correctness (Outcome) shrinks the median crop area to 1.00\%, leaves 54.44\% of correct answers ungrounded ($U'$), adds only $+0.04$ Avg$_6$ points from views, and recovers just 30.12\% of the GT-crop gap ($+6.86$ points behind GT crops).
By contrast, PEA's overlap reward produces well-proportioned crops (3.88\% area), lifts hit rate and IoU to 78.59\% and 0.68 (vs.\ 73.98\% and 0.53 for VPD without BAT), cuts $U'$ to 18.28\%, boosts the view gain to $+1.83$ Avg$_6$ points, and recovers 101.37\% of the gap to ground-truth crops.

\input{fig/figS_cost.tex}
\input{tab/supp_tool_loops.tex}

\subsection{Zooming and Inference Cost}
\label{app:inference-cost}

To test whether LEAP's lead comes from its training rather than test-time zooming alone, Fig.~\ref{fig:cost} and Table~\ref{tab:tool-loops} compare three modes across the four perception benchmarks (V$^\star$, ZoomBench, and HR-Bench 4K/8K; 2,636 questions): single pass; zooming, where the zoom tool is available and the model decides when to call it; and forced zooming, where a released student's first call is forced.
No released student reaches LEAP in any mode: every student peaks in its single pass, which LEAP still leads by $2.44$ to $6.61$ points.
Whereas zooming lifts the base model ($+4.22$ points) and LEAP ($+3.01$ $[1.55,4.47]$ points), autonomous zooming ($1.41\times$ to $2.54\times$ GPU time) drops the released students by $2.28$ to $7.40$ points below their single pass, and forced zooming ($2.10\times$ to $2.84\times$ GPU time) still leaves them $0.91$ to $3.17$ points below single pass.
Passive distillation therefore damages multi-turn view utilization as well as tool invocation.
LEAP's zooming (85.19\%) leads the students' autonomous zooming by $+4.73$ $[2.72,6.80]$ to $+14.01$ $[10.69,17.30]$ points and their forced zooming by $+3.36$ $[0.80,5.74]$ to $+9.77$ $[6.38,12.99]$ points (closest: OPD-Aha forced zooming at 81.83\% and $2.49\times$ GPU time, vs.\ LEAP's $2.34\times$).

All models are timed on a single Nvidia GB200 GPU with the same serving stack and official judge, measuring wall-clock latency at concurrency 1 (400 questions, 100 per benchmark) and GPU-seconds at concurrency 64 (381 questions) relative to a Qwen3.5-4B single pass.
At $3.32\times$ latency and $2.34\times$ GPU time of the base single pass, LEAP outperforms the Thinking-with-Images RL agents (with SenseNova-MARS-8B restricted to its crop tool) by $7.32$ to $17.01$ points while taking lower latency than DeepEyes and DeepEyesV2 ($1.41\times$ to $1.98\times$ LEAP's latency at $0.79\times$ to $0.95\times$ its GPU time).

\subsection{Training Seeds}
\label{app:seeds}
\input{tab/supp_seeds.tex}

Table~\ref{tab:seeds} evaluates LEAP-4B and LEAP-9B across four random training seeds (seed 42 is the model of Table~\ref{tab:main}; Appendix~\ref{app:leap_impl} gives the seed of every other row).
At 4B, Avg$_6$ reaches $82.29\pm0.19\%$ ($82.15\%$ to $82.56\%$), so every seed exceeds OPD-Aha-4B ($80.54\%$) by at least $1.61$ Avg$_6$ points and by at least $5.92$ points on ZoomBench ($68.76\%$ vs.\ $62.84\%$).
At 9B, Avg$_6$ reaches $82.68\pm0.09\%$ ($82.56\%$ to $82.77\%$), so every seed exceeds every 9B baseline by at least $0.78$ Avg$_6$ points (OPD-Aha-9B, $81.78\%$) and $2.84$ points on ZoomBench (Vision-OPD-9B, $65.56\%$).

\input{fig/figS_dynamics.tex}
\subsection{Training Dynamics}
\label{app:dynamics}

Fig.~\ref{fig:dynamics} tracks the three factors of Eq.~\eqref{eq:leap_policy} across training; for the released students, panel (a) uses their Table~\ref{tab:main} accuracy, while (b) and (c) follow the same acquisition procedure.
In (a), ZoomBench accuracy rises steadily from 58.11\% for the base model to 62.72\%, 66.51\%, 68.28\%, and 69.59\% across LEAP's four checkpoints, surpassing all four released students (59.76\%--62.84\%) by the second checkpoint and finishing $+3.31$ $[0.83,5.80]$ points above VPD (Appendix~\ref{app:intervals}).
In (b), LEAP's views tighten around the evidence from 0.57 to 0.68 IoU, whereas passive distillation leaves VPD (0.53), the crop teacher (0.52), and the released students (0.45--0.53) below the base model.
In (c), LEAP's inspection AUROC rises from 0.56 to 0.62, whereas VPD drops to 0.39, the crop teacher to 0.44, and the released students range from 0.44 to 0.47.

%% file: tab/supp1_bias.tex
\begin{table}[t]
\centering
\caption{\textbf{LEAP affirms fewer false claims than every released student.} Base: levels (\%); others: changes from base (pp). Forced answer readouts; HB: HallusionBench (Appendix~\ref{app:diagnostics}).}
\label{tab:bias}
\begingroup
\tabsetup{10}{13}{5pt}
\begin{tabularx}{\linewidth}{l *{5}{C}}
\toprule
& \multicolumn{2}{c}{Says Yes} & Net claim & HB false & Masked \\
\cmidrule(lr){2-3}
Model & True claim$\uparrow$ & False claim$\downarrow$ & verification$\uparrow$ & alarm$\downarrow$ & abstention$\uparrow$ \\
\midrule
Base (levels, \%) & 67.1 & 35.1 & 66.0 & 29.4 & 74.0 \\
Vision-OPD & +16.0 & +17.4 & \textminus0.7 & +14.0 & \textminus25.0 \\
VAD & +21.2 & +25.9 & \textminus2.3 & +18.1 & \textminus45.0 \\
OPD-Aha & +13.9 & +14.1 & \textminus0.1 & +10.1 & \textminus45.5 \\
OPD-V & +23.8 & +27.7 & \textminus2.0 & +20.1 & \textminus43.5 \\
\leaprow \textbf{LEAP} & +13.5 & +10.2 & +1.7 & +7.0 & \textminus35.5 \\
\midrule
\tabgroup{6}{Same pipeline (Table~\ref{tab:components} checkpoints)}
Crop teacher & +10.4 & +8.6 & +0.9 & +8.7 & \textminus15.5 \\
VPD & +14.5 & +12.5 & +1.0 & +10.0 & \textminus34.5 \\
VPD + PEA & +12.9 & +9.9 & +1.5 & +6.7 & \textminus38.5 \\
VPD + BAT & +15.3 & +13.1 & +1.1 & +9.3 & \textminus33.0 \\
Crop teacher + BAT + PEA & +9.0 & +5.6 & +1.7 & +8.3 & \textminus18.5 \\
\bottomrule
\end{tabularx}
\endgroup
\end{table}

%% file: tab/supp2_native.tex
\begin{table}[t]
\centering
\caption{\textbf{Passive distillation suppresses native zoom calls.} Qwen3.5-4B with the native zoom tool and official judge. Bold: best among released students and LEAP (Appendix~\ref{app:diagnostics}).}
\label{tab:native}\label{tab:native-decisions}
\begingroup
\tabsetup{9}{11.6}{3.5pt}
\begin{tabularx}{\linewidth}{L *{7}{c}}
\toprule
& \multicolumn{2}{c}{Zoom calls (\%)} & Forced & \multicolumn{2}{c}{When calling (\%)} & \multicolumn{2}{c}{Accuracy change (pp)} \\
\cmidrule(lr){2-3}\cmidrule(lr){5-6}\cmidrule(l){7-8}
Model & All$\uparrow$ & Own errors$\uparrow$ & hit (\%)$\uparrow$ & Fix$\uparrow$ & Break$\downarrow$ & Recovery$\uparrow$ & Model decides$\uparrow$ \\
\midrule
Base & 100.00 & 100.00 & 75.65 & 47.5 & 12.5 & \tabna & +9.47 \\
Vision-OPD & 49.74 & 58.59 & 74.34 & 31.4 & 19.8 & \textminus13.78 & \textminus6.28 \\
VAD & 58.67 & 66.79 & \textbf{76.27} & 43.3 & 18.7 & \textminus5.84 & \textminus2.09 \\
OPD-Aha & 22.75 & 19.97 & 74.15 & \textbf{43.5} & 12.6 & \textminus17.97 & \textminus2.37 \\
OPD-V & 7.11 & 8.75 & 74.02 & 31.3 & 32.6 & \textminus19.14 & \textminus7.09 \\
\leaprow \textbf{LEAP} & \textbf{87.26} & \textbf{87.95} & 75.24 & 41.2 & \textbf{9.2} & \textbf{\textminus1.94} & \textbf{+0.76} \\
\midrule
\tabgroup{8}{Same pipeline (Table~\ref{tab:components} checkpoints)}
Crop teacher & 48.28 & 56.19 & 75.61 & 37.1 & 18.8 & \textminus10.81 & \textminus7.12 \\
VPD & 12.96 & 16.80 & 79.85 & 39.2 & 12.4 & \textminus21.84 & \textminus2.52 \\
VPD + PEA & 33.21 & 32.45 & 73.07 & 18.4 & 39.1 & \textminus22.60 & \textminus6.68 \\
VPD + BAT & 1.87 & 3.47 & 79.16 & 50.0 & 11.1 & \textminus24.75 & \textminus3.96 \\
Crop teacher + BAT + PEA & 97.58 & 96.76 & 71.77 & 39.3 & 15.0 & \textminus2.96 & +1.55 \\
\bottomrule
\end{tabularx}
\endgroup
\end{table}

%% file: tab/supp_ablation_full.tex
\begin{table}[t]
\centering
\caption{\textbf{Complete six-benchmark ablations.} Official accuracy (\%); Qwen3.5-4B. Every row, Base included, uses the acquisition procedure with its own held-out threshold, whereas Table~\ref{tab:main} evaluates the base checkpoint directly.}
\label{tab:ablation-full}
\begingroup
\tabsetup{9.2}{11.8}{4pt}
\begin{tabularx}{\linewidth}{L *{6}{c} A}
\toprule
Training & V$^{\star}$ & ZoomBench & \shortstack{HR-Bench\\4K} & \shortstack{HR-Bench\\8K} & \shortstack{MME-RW\\EN} & \shortstack{MME-RW\\CN} & \textbf{Avg$_6$} \\
\midrule
\tabgroupavg{7}{Components}
Base & 89.01 & 58.11 & 85.38 & 82.12 & 67.31 & 70.39 & 75.39 \\
Crop teacher, $\beta=0$ & 91.10 & 67.46 & 85.50 & 82.75 & 74.46 & 70.93 & 78.70 \\
VPD & 93.72 & 66.27 & 90.12 & 88.00 & 78.23 & 74.90 & 81.88 \\
VPD + PEA & 93.72 & 67.34 & 89.62 & 88.88 & 77.93 & 74.16 & 81.94 \\
VPD + BAT & 90.58 & 67.57 & 90.12 & 88.62 & 78.08 & 74.73 & 81.62 \\
\tabgroupavg{7}{Privileged target}
LEAP, $\beta=0$ & 86.39 & 66.86 & 84.00 & 84.38 & 73.35 & 70.20 & 77.53 \\
LEAP, $\beta=2$ & 90.58 & 68.17 & 90.25 & 88.00 & 77.82 & 73.45 & 81.38 \\
LEAP, degraded-crop contrast & 93.19 & 66.98 & 89.75 & 86.50 & 76.99 & 71.96 & 80.90 \\
\tabgroupavg{7}{Divergence and region reward}
Forward KL, $\beta=0$ & 87.96 & 65.68 & 83.25 & 83.12 & 74.12 & 70.31 & 77.41 \\
Reverse KL, $\beta=0$ & 88.48 & 67.34 & 83.12 & 83.75 & 73.57 & 69.77 & 77.67 \\
PEA with outcome reward & 89.01 & 64.50 & 89.25 & 87.38 & 77.57 & 73.74 & 80.24 \\
\midrule
\leaprow \textbf{LEAP} & 92.67 & 69.59 & 90.00 & 88.50 & 78.30 & 74.40 & 82.24 \\
\bottomrule
\end{tabularx}
\endgroup
\end{table}

%% file: tab/supp_target_choice.tex
\begin{table}[t]
\centering
\caption{\textbf{Choosing the privileged target.} JSD distillation with acquisition training; Qwen3.5-4B.}
\label{tab:hyperparameters}
\begingroup
\tabsetup{10}{13}{6pt}
\begin{tabularx}{\linewidth}{l *{3}{C}}
\toprule
Contrast view & $\beta$ & ZoomBench$\uparrow$ & Avg$_6\uparrow$ \\
\midrule
None & 0 & 66.86 & 77.53 \\
Null view & 2 & 68.17 & 81.38 \\
Degraded crop & 4 & 66.98 & 80.90 \\
\leaprow \textbf{Null view} & 4 & \textbf{69.59} & \textbf{82.24} \\
\bottomrule
\end{tabularx}
\endgroup
\end{table}

%% file: tab/supp_acquisition_interaction.tex
\begin{table}[t]
\centering
\caption{\textbf{Interaction between target contrast and acquisition training.} Avg$_6$ (\%) without and with acquired views; gains (pp) on common valid items.}
\label{tab:acquisition-interaction}
\begingroup
\tabsetup{10}{13}{6pt}
\begin{tabularx}{\linewidth}{l *{3}{C}}
\toprule
Training & Direct & With views & Gain (pp) \\
\midrule
$\beta=0$, acquisition training off & 77.13 & 78.70 & +1.57 \\
$\beta=0$, acquisition training on & 77.59 & 77.53 & \textminus0.06 \\
$\beta=4$, acquisition training off & 80.59 & 81.88 & +1.28 \\
\leaprow \textbf{\boldmath$\beta=4$, acquisition training on} & 80.42 & 82.24 & \textbf{+1.83} \\
\bottomrule
\end{tabularx}
\endgroup
\end{table}

%% file: tab/supp_ablation_bias.tex
\begin{table}[t]
\centering
\caption{\textbf{Answer bias across distillation settings.} False affirmation (\%) under the controlled readout on the 642 gold-No HallusionBench questions (Original) and the false claims of the 815 original/reversed pairs (Paired).}
\label{tab:ablation-bias}
\begingroup
\tabsetup{10}{13}{6pt}
\begin{tabularx}{\linewidth}{l *{4}{C}}
\toprule
Training & $\beta$ & Avg$_6\uparrow$ & Original$\downarrow$ & Paired$\downarrow$ \\
\midrule
Base & \tabna & 75.39 & 29.44 & 35.09 \\
JSD & 0 & 77.53 & 37.69 & 40.74 \\
JSD & 2 & 81.38 & 39.56 & 46.99 \\
\leaprow \textbf{JSD (LEAP)} & 4 & 82.24 & 36.45 & 45.28 \\
JSD, degraded crop & 4 & 80.90 & 41.59 & 48.59 \\
Forward KL & 0 & 77.41 & 40.34 & 46.01 \\
Reverse KL & 0 & 77.67 & 35.05 & 39.88 \\
\bottomrule
\end{tabularx}
\endgroup
\end{table}

%% file: fig/figS_when.tex
\begin{figure}[t]
\centering
\includegraphics[width=\linewidth]{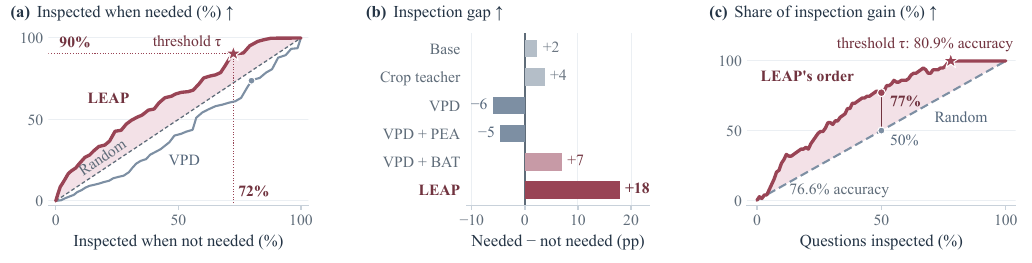}
\caption{\textbf{With BAT, a fixed threshold inspects the questions that need inspection more often.} Controlled experiment on questions collected post hoc that need or do not need inspection. (a) Inspection rates of the two groups as the threshold varies (markers: calibrated threshold $\tau$; dashed: random selection). (b) Gap between the two rates at $\tau$ for the trainings of Table~\ref{tab:bat}. (c) Share of the accuracy gain from inspection obtained when LEAP inspects in the order of its inspection score, against random selection (star: $\tau$).}
\label{fig:when-to-look}
\end{figure}

%% file: tab/supp_inspection_priority.tex
\begin{table}[t]
\centering
\caption{\textbf{BAT lifts the error ranking and the fix ranking above chance.} Additional metrics for Table~\ref{tab:bat}; Recall: error recall (\%) at a 72.6\% inspection rate.}
\label{tab:behaviour-details}
\begingroup
\tabsetup{10}{13}{6pt}
\begin{tabularx}{\linewidth}{l *{3}{C}}
\toprule
Training & AUROC$\uparrow$ & Recall$\uparrow$ & AUROC$_{\mathrm{fix}}\uparrow$ \\
\midrule
Base & 0.56 & 74.69 & 0.43 \\
Crop teacher & 0.44 & 51.99 & 0.37 \\
VPD & 0.39 & 50.42 & 0.32 \\
VPD + PEA & 0.38 & 52.38 & 0.35 \\
VPD + BAT & 0.57 & 79.69 & 0.55 \\
\leaprow \textbf{LEAP} & \textbf{0.62} & \textbf{83.33} & \textbf{0.60} \\
\bottomrule
\end{tabularx}
\endgroup
\end{table}

%% file: tab/supp_behaviour_details.tex
\begin{table}[t]
\centering
\caption{\textbf{With PEA's overlap reward, acquired views match ground-truth crops on ZoomBench.} Additional metrics for Table~\ref{tab:pea}; ZoomBench except Gain (Avg$_6$), VPD active. Views, Area, Closed: \%; Gain, GT gain: pp (Appendix~\ref{app:region-selection}).}
\label{tab:grounding-details}
\begingroup
\tabsetup{9.6}{12.4}{4pt}
\begin{tabularx}{\linewidth}{l *{8}{C}}
\toprule
PEA reward & Views & Area & Hit$\uparrow$ & IoU$\uparrow$ & $U'\downarrow$ & Gain & Closed & GT gain \\
\midrule
None (VPD + BAT) & 93.37 & 3.50 & 74.52 & 0.54 & 22.05 & 0.70 & \tabna & \tabna \\
Outcome & 93.25 & 1.00 & 42.51 & 0.23 & 54.44 & 0.04 & 30.12 & +6.86 \\
IoU (VPD + PEA) & 88.17 & 3.28 & \textbf{79.73} & \textbf{0.68} & 19.32 & 1.67 & \tabna & \tabna \\
\leaprow \textbf{IoU (LEAP)} & 93.96 & 3.88 & 78.59 & \textbf{0.68} & \textbf{18.28} & \textbf{1.83} & \textbf{101.37} & \textminus0.12 \\
\bottomrule
\end{tabularx}
\endgroup
\end{table}

%% file: fig/figS_cost.tex
\begin{figure}[t]
\centering
\includegraphics[width=\linewidth]{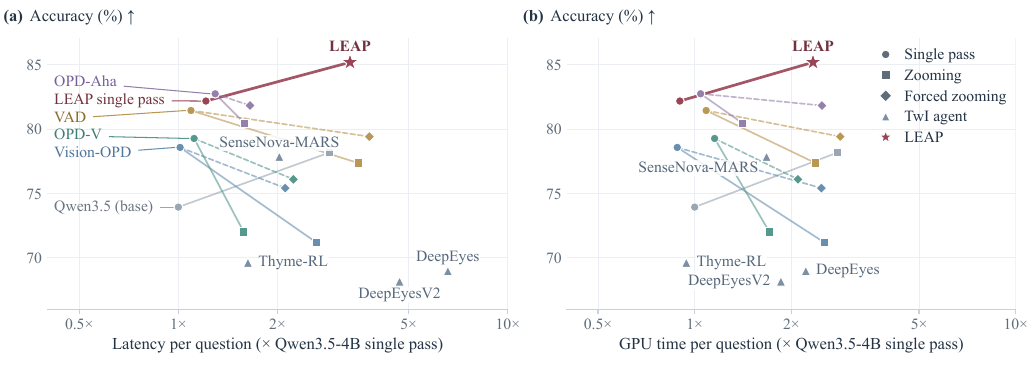}
\caption{\textbf{No released student reaches LEAP in any mode: zooming lowers student accuracy, whereas the base and LEAP gain.} To test whether zooming at test time brings the released 4B students to LEAP, we plot accuracy on the four perception benchmarks (full sets, Table~\ref{tab:tool-loops}) against (a) latency and (b) GPU time per question, both relative to a Qwen3.5-4B single pass (log scale). Thin lines lead from each model's single pass to its zooming (solid) and forced zooming (dashed); the thick line leads to LEAP's zooming.}
\label{fig:cost}
\end{figure}

%% file: tab/supp_tool_loops.tex
\begin{table}[t]
\centering
\caption{\textbf{Zooming at test time does not bring the released students to LEAP.} Accuracy (\%) of Qwen3.5-4B, the four released 4B students, and LEAP-4B on the four perception benchmarks (full sets, Fig.~\ref{fig:cost}). Single pass: no tool; Zooming: the tool is available and the model decides whether to call it; Forced zooming: a student's first call is forced; Three-sample vote: the majority answer of three single passes sampled at temperature 1, at a token budget matched to LEAP's zooming; $\Delta$: Avg$_4$ change from single pass (pp); latency and GPU time are relative to the base single pass.}
\label{tab:tool-loops}
\begingroup
\tabsetup{9.2}{11.6}{3.5pt}
\begin{tabularx}{\linewidth}{L l *{4}{c} A c c c}
\toprule
Model & Mode & V$^{\star}$ & ZoomBench & \shortstack{HR-Bench\\4K} & \shortstack{HR-Bench\\8K} & \textbf{Avg$_4$} & $\Delta$ & Latency & \shortstack{GPU\\time} \\
\midrule
Base & Single pass & 81.15 & 48.76 & 85.62 & 80.25 & 73.95 & \tabna & 1.00 & 1.00 \\
 & Zooming & 84.82 & 64.38 & 82.00 & 81.50 & 78.17 & +4.22 & 2.87 & 2.78 \\
\midrule
Vision-OPD & Single pass & 90.05 & 59.76 & 83.25 & 81.25 & 78.58 & \tabna & 1.01 & 0.88 \\
 & Zooming & 80.63 & 56.45 & 75.50 & 72.12 & 71.18 & \textminus7.40 & 2.63 & 2.54 \\
 & Forced zooming & 85.34 & 62.60 & 78.25 & 75.50 & 75.42 & \textminus3.16 & 2.11 & 2.48 \\
\midrule
VAD & Single pass & 93.72 & 60.95 & 87.25 & 83.88 & 81.45 & \tabna & 1.09 & 1.09 \\
 & Zooming & 89.53 & 60.95 & 80.38 & 78.62 & 77.37 & \textminus4.08 & 3.52 & 2.38 \\
 & Forced zooming & 90.05 & 66.75 & 79.38 & 81.50 & 79.42 & \textminus2.03 & 3.80 & 2.84 \\
\midrule
OPD-Aha & Single pass & \textbf{94.76} & 62.84 & 88.12 & 85.25 & 82.74 & \tabna & 1.29 & 1.04 \\
 & Zooming & 90.58 & 61.89 & 84.62 & 84.75 & 80.46 & \textminus2.28 & 1.59 & 1.41 \\
 & Forced zooming & 94.24 & 65.09 & 85.00 & 83.00 & 81.83 & \textminus0.91 & 1.65 & 2.49 \\
\midrule
OPD-V & Single pass & 90.05 & 61.42 & 84.12 & 81.50 & 79.27 & \tabna & 1.12 & 1.15 \\
 & Zooming & 81.68 & 55.98 & 75.25 & 75.12 & 72.01 & \textminus7.26 & 1.58 & 1.71 \\
 & Forced zooming & 87.96 & 60.83 & 78.25 & 77.38 & 76.10 & \textminus3.17 & 2.23 & 2.10 \\
\midrule
\leaprow \textbf{LEAP} & Single pass & 92.15 & 60.83 & 89.38 & 86.38 & 82.18 & \tabna & 1.21 & 0.90 \\
\leaprow  & Three-sample vote & 93.19 & 61.18 & 88.62 & 86.50 & 82.38 & +0.20 & \tabna & \tabna \\
\leaprow  & Zooming & 92.67 & \textbf{69.59} & \textbf{90.00} & \textbf{88.50} & \textbf{85.19} & +3.01 & 3.32 & 2.34 \\
\bottomrule
\end{tabularx}
\endgroup
\end{table}

%% file: tab/supp_seeds.tex
\begin{table}[t]
\centering
\caption{\textbf{LEAP across training seeds.} Accuracy (\%) on the six benchmarks of Table~\ref{tab:main}; Mean and Std are taken over the four seeds.}
\label{tab:seeds}
\begingroup
\tabsetup{8.8}{11.2}{2.6pt}\setlength{\aboverulesep}{0.25ex}\setlength{\belowrulesep}{0.45ex}
\begin{tabularx}{\linewidth}{l >{\centering\arraybackslash}X c *{4}{>{\centering\arraybackslash}X} >{\columncolor{leapavg}\centering\arraybackslash}X @{\hspace{10pt}} >{\centering\arraybackslash}X c *{4}{>{\centering\arraybackslash}X} >{\columncolor{leapavg}\centering\arraybackslash}X}
\toprule
 & \multicolumn{7}{c}{LEAP-4B} & \multicolumn{7}{c}{LEAP-9B} \\
\cmidrule(lr{10pt}){2-8}\cmidrule(l{0pt}r{0pt}){9-15}
 & & & \multicolumn{2}{c}{HR-Bench} & \multicolumn{2}{c}{MME-RW} & \multicolumn{1}{c}{} & & & \multicolumn{2}{c}{HR-Bench} & \multicolumn{2}{c}{MME-RW} & \multicolumn{1}{c}{} \\
Seed & V$^{\star}$ & ZoomBench & 4K & 8K & EN & CN & \textbf{Avg$_6$} & V$^{\star}$ & ZoomBench & 4K & 8K & EN & CN & \textbf{Avg$_6$} \\
\midrule
42 & 92.67 & 69.59 & 90.00 & 88.50 & 78.30 & 74.40 & 82.24 & 94.76 & 69.70 & 90.75 & 90.25 & 77.38 & 72.49 & 82.56 \\
43 & 93.72 & 69.82 & 89.38 & 89.50 & 78.09 & 74.85 & 82.56 & 94.24 & 69.23 & 89.88 & 90.88 & 77.68 & 74.23 & 82.69 \\
44 & 91.10 & 71.60 & 88.50 & 89.38 & 78.47 & 74.18 & 82.20 & 93.19 & 68.40 & 91.75 & 90.12 & 78.44 & 74.68 & 82.77 \\
3407 & 94.24 & 68.76 & 89.00 & 88.12 & 78.66 & 74.14 & 82.15 & 95.29 & 70.18 & 89.88 & 90.75 & 76.41 & 73.75 & 82.71 \\
\midrule
Mean & 92.93 & 69.94 & 89.22 & 88.88 & 78.38 & 74.39 & 82.29 & 94.37 & 69.38 & 90.56 & 90.50 & 77.47 & 73.79 & 82.68 \\
Std & 1.39 & 1.20 & 0.63 & 0.67 & 0.24 & 0.33 & 0.19 & 0.89 & 0.76 & 0.89 & 0.37 & 0.84 & 0.95 & 0.09 \\
\bottomrule
\end{tabularx}
\endgroup
\end{table}

%% file: fig/figS_dynamics.tex
\begin{figure}[t]
\centering
\includegraphics[width=\linewidth]{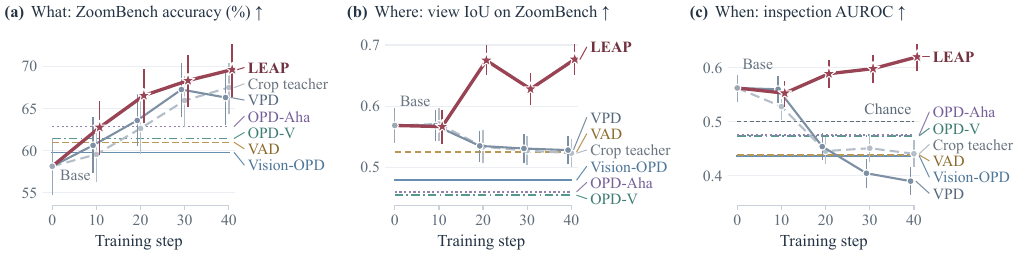}
\caption{\textbf{Training dynamics, LEAP learns what to answer, where to look, and when to look.} Horizontal lines mark the four released passive students. (a) ZoomBench accuracy with test-time acquisition. (b) IoU between acquired views and the evidence region on ZoomBench. (c) Inspection AUROC on the 2,546-question audit set. Whiskers: 95\% intervals.}
\label{fig:dynamics}
\end{figure}

%% file: sec/A0_evaluation_protocol.tex
\section{Benchmarks and Evaluation Protocol}
\label{app:evaluation-details}

This section details the benchmarks, the shared scoring protocol, and the compared methods of Sec.~\ref{sec:experimental-setup}; Appendix~\ref{app:remote-sensing} covers the remote-sensing transfer.

\subsection{Benchmarks}
\label{app:benchmarks}
\input{tab/supp_benchmarks.tex}

Table~\ref{tab:benchmarks} lists the evaluation sets and their composition.
The six fine-grained benchmarks hold 32,162 questions, all of which we use.
Vision-OPD's data script~\citep{yuan2026visionopd} turns each benchmark into one question file that fixes the image, the query text, and the gold answer of every question for all models.
These queries append each benchmark's answer instruction to the question (Appendix~\ref{app:prompts}).

\subsection{Evaluation Protocol}
\label{app:eval-protocol}

\paragraph{Requests.}
The single-pass answers of the released checkpoints, the base models, and the API models use the request of Vision-OPD's inference script: one user turn with the image and the query, no system prompt, and temperature 0, with thinking disabled for the Qwen3.5-based models.
V$^{\star}$ images are re-encoded as PNG; the other images are sent as stored.
LEAP answers through its test-time acquisition procedure (Appendix~\ref{app:impl-inference}) and is scored like every other row.

\paragraph{Answer extraction and judging.}
We run Vision-OPD's judge script unchanged, with gpt-oss-120b~\citep{openai2025gptoss} as the judge model.
The script first extracts the answer: the text inside \texttt{<answer>} tags if present, otherwise the text from the first ``Answer:'' onward, otherwise the whole response.
A rule stage then accepts the response if the \texttt{mathruler} grader matches it to the gold answer or, on the multiple-choice V$^{\star}$, HR-Bench, and MME-RealWorld, if the first option letter in it equals the gold letter.
Every other response goes to the judge, which receives the question, the gold answer, and the extracted response in Vision-OPD's prompt (Appendix~\ref{app:prompts}), decodes at temperature 0 with at most 2,048 new tokens, and counts the response as correct when it replies Yes.

\paragraph{Metrics.}
The accuracy of a benchmark is the share of its questions judged correct, with failed requests counted as wrong.
Avg$_6$ is the unweighted mean of the six benchmark accuracies, and the four-benchmark perception mean in Sec.~\ref{sec:main-results} averages V$^{\star}$, ZoomBench, and both HR-Bench resolutions.
While Table~\ref{tab:main} uses Vision-OPD's unchanged judge for direct comparability, the remote-sensing benchmarks add an option-conflict check (Appendix~\ref{app:remote-sensing}).
Applying that check to Table~\ref{tab:main} yields 80.71\% Avg$_6$ for LEAP-4B versus 78.58\%, 77.99\%, 75.57\%, and 75.23\% for OPD-Aha, VAD, OPD-V, and Vision-OPD at 4B, preserving their exact ranking while widening LEAP-4B's lead over OPD-Aha from $+1.70$ to $+2.13$ points.

\subsection{Compared Methods}
\label{app:baselines}
\input{tab/supp_tools_off.tex}

Published baselines identified in Sec.~\ref{sec:experimental-setup} use their reported averages, and Avg$_6$ of ZwZ2~\citep{wei2026zwz2} is the mean of its six reported accuracies; all other rows are scored with the protocol above.
The released Qwen3.5-based checkpoints receive the single-turn request above.
SFT is our supervised fine-tuning of Qwen3.5-4B and Qwen3.5-9B on the reference answers of Vision-OPD-6K~\citep{yuan2026visionopd}.

\paragraph{API models.}
API models receive the same request with the same image bytes through OpenRouter, pinned to one provider without fallback;
Gemini 3.8 Flash, Claude Haiku 5.5, and GPT-6-luna use provider-default reasoning, whereas Qwen3.5-397B-A17B, GLM-4.6V, Kimi K3, and DeepSeek v4.1 Flash run with reasoning turned off.

\paragraph{``Thinking with Images'' agents.}
We evaluate all four methods' visual tool loops across all 32,162 questions of the six benchmarks using their official templates and decoding settings, with at most 11 zoom calls (DeepEyes), 11 Python generations (DeepEyesV2), five attempts of five code-execution iterations (Thyme-RL), and 50 turns (SenseNova-MARS).
DeepEyes boxes are mapped from the processed view to the original image before cropping; Thyme-RL re-encodes returned images with the conversation and uses the first five MME-RealWorld options per its released prompt.
Disabling their tool loops (Table~\ref{tab:tools-off}) lowers their Avg$_6$ by $0.43$ to $5.82$ points (e.g., SenseNova-MARS drops from $75.49\%$ to $69.67\%$), so Table~\ref{tab:main} reports each agent in its stronger setting, which LEAP-4B leads by $6.75$ to $15.45$ points.

%% file: tab/supp_benchmarks.tex
\begin{table}[t]
\centering
\caption{\textbf{Evaluation benchmarks.} Every set is used in full. MC: multiple choice, answered with an option letter.}
\label{tab:benchmarks}
\begingroup
\tabsetup{10}{13}{6pt}
\begin{tabularx}{\linewidth}{l r X}
\toprule
Benchmark & Questions & Answer format and composition \\
\midrule
\tabgroup{3}{Fine-grained perception (Table~\ref{tab:main})}
V$^{\star}$~\citep{wu2024vstar} & 191 & MC; 115 attribute and 76 position questions \\
ZoomBench~\citep{wei2026zwz} & 845 & 621 MC and 224 counting questions \\
HR-Bench 4K~\citep{wang2025hrbench} & 800 & MC (A--D); 400 single-, 400 cross-instance \\
HR-Bench 8K~\citep{wang2025hrbench} & 800 & MC (A--D); 400 single-, 400 cross-instance \\
MME-RealWorld EN~\citep{zhang2025mmerealworld} & 23,609 & MC (A--E); English \\
MME-RealWorld CN~\citep{zhang2025mmerealworld} & 5,917 & MC (A--E); Chinese \\
\midrule
\tabgroup{3}{Remote-sensing transfer (Table~\ref{tab:remote-sensing})}
XLRS-Bench-lite~\citep{wang2025xlrsbench} & 3,080 & MC; 13 subtasks \\
MME-RealWorld-RS~\citep{zhang2025mmerealworld} & 3,738 & MC (A--E); color, counting, position \\
LRS-VQA~\citep{luo2025lrsvqa} & 7,333 & short phrase; 8 categories, 3 sources \\
\bottomrule
\end{tabularx}
\endgroup
\end{table}

%% file: tab/supp_tools_off.tex
\begin{table}[t]
\centering
\caption{\textbf{``Thinking with Images'' agents evaluated with tools disabled.} Official single-turn accuracy without tool loops; Table~\ref{tab:main} evaluates each agent in its stronger tool-enabled mode.}
\label{tab:tools-off}
\begingroup
\tabsetup{9.6}{12.4}{4.5pt}
\begin{tabularx}{\linewidth}{L c *{6}{c} A}
\toprule
Model & Size & V$^{\star}$ & ZoomBench & \shortstack{HR-Bench\\4K} & \shortstack{HR-Bench\\8K} & \shortstack{MME-RW\\EN} & \shortstack{MME-RW\\CN} & \textbf{Avg$_6$} \\
\midrule
DeepEyes & 7B & 80.10 & 46.27 & 75.25 & 70.50 & 61.86 & 59.24 & 65.54 \\
DeepEyesV2 & 7B & 80.63 & 46.63 & 76.62 & 71.50 & 60.16 & 62.62 & 66.36 \\
SenseNova-MARS & 8B & 83.77 & 44.02 & 81.12 & 77.38 & 65.03 & 66.67 & 69.67 \\
Thyme-RL & 7B & 81.68 & 46.86 & 77.25 & 69.25 & 57.66 & 57.31 & 65.00 \\
\bottomrule
\end{tabularx}
\endgroup
\end{table}

%% file: sec/B_remote_sensing_protocol.tex
\section{Remote-Sensing Evaluation}
\label{app:remote-sensing}
This section details the evaluation protocol for the remote-sensing stress testbed in Sec.~\ref{sec:remote-sensing} (Table~\ref{tab:remote-sensing}), introduced in Sec.~\ref{sec:experimental-setup}.

\paragraph{Benchmarks and inputs.}
To test out-of-domain generalization under extreme resolution asymmetry, Table~\ref{tab:remote-sensing} evaluates the full XLRS-Bench-lite (3,080 questions across 13 subtasks), MME-RealWorld-RS (3,738 questions across color, counting, and position), and LRS-VQA (7,333 questions averaged across categories within each of three sources, then across sources) benchmarks (Table~\ref{tab:benchmarks}).
Following RS-OPSD~\citep{jiang2026rsopsd}, XLRS-Bench-lite averages its 13 subtask accuracies and MME-RealWorld-RS its three category accuracies; Avg is the unweighted mean of the three benchmarks.
All models share the benchmark questions and full-image preprocessing of the released RS-OPSD client while running their own inference procedures.
The client resizes each image with Lanczos resampling to at most 16,777,216 pixels, sends it as PNG, and decodes at temperature 0 with thinking disabled (4,096-token limit).

\paragraph{Shared corrected judge.}
To prevent rule-matching artifacts when models mention multiple options before concluding, our shared corrected judge runs the Vision-OPD judge~\citep{yuan2026visionopd} with gpt-oss-120b~\citep{openai2025gptoss} and rejudges rule-accepted responses whose final option conflicts with the gold option.
XLRS-Bench-lite and MME-RealWorld-RS take the multiple-choice rules of Appendix~\ref{app:eval-protocol}, and the free-form LRS-VQA takes the same path as ZoomBench.

\paragraph{Baselines.}
We evaluate RS-OPSD and RS-OPD-Lite from their released checkpoints under the same client and shared corrected judge (RS-OPD-Lite distills an 8B teacher into a 2B student; Table~\ref{tab:remote-sensing} lists student size).
Under the released final-answer scorer, our RS-OPSD-8B evaluation matches its published average within 0.7 points.

\paragraph{Zero-shot and in-domain LEAP.}
For zero-shot transfer, LEAP-4B and LEAP-9B keep their Vision-OPD-6K weights without any remote-sensing training; only the BAT trigger threshold is set from unlabeled decision scores on 480 GeoEvidence examples held out by source image (targeting a 90\% inspection share without answer labels).
For in-domain adaptation, LEAP-RS initializes from Qwen3.5-4B and trains the full LEAP objective on the single-answer portion of GeoEvidence-6K~\citep{jiang2026rsopsd}, excluding the 480 held-out calibration images, filtering truncated distillation responses under a 2,048-token training limit.
On remote sensing, images above the processor cap are localized on their downscaled view, views are cut from the original full-resolution image, and answer turns use the client's query and 4,096-token limit.
On XLRS-Bench-lite, whose object questions state their target box in original-image pixels, a predicted box that is invalid on the 0--1000 grid but fits the original image is read in those pixel coordinates.

%% file: sec/D_implementation.tex
\section{Implementation Details}
\label{app:leap_impl}
\input{tab/supp_train_settings.tex}

This section gives the implementation details behind Secs.~\ref{sec:leap_training} and~\ref{sec:experimental-setup}; Table~\ref{tab:train-settings} summarizes the training settings, and Appendix~\ref{app:prompts} gives the prompt templates.

\subsection{Training Inputs and Shared Policy}
\label{app:impl-inputs}

The three losses supervise the three token segments of the single shared policy $\Pi_\theta$ in Eq.~\eqref{eq:leap_policy}: $\mathcal L_{\mathrm{BAT}}$ the first-token trigger $g_\theta(z\mid x)$, $\mathcal L_{\mathrm{PEA}}$ the tool-call distribution $q_\theta(u\mid x,1)$, and $\mathcal L_{\mathrm{VPD}}$ the answer tokens on the region-marked image, with a teacher $\pi_\phi$ that is a frozen copy of the initial backbone.
Each training example provides an image $I$, a question $q$, a reference answer $a^\star$, and a privileged bounding box $b^\star$.
We train on Vision-OPD-6K~\citep{yuan2026visionopd} in a fixed order, visiting each training example at most once; the 480 held-out examples serve only to calibrate the inference threshold (Appendix~\ref{app:impl-inference}).
To train all three token stages in parallel without waiting for localization to converge, VPD supplies $\pi_\theta$ with the region-marked full image $I^g$ (instructing it to focus on the red box) while the teacher observes the crop $I^\star=\operatorname{Crop}(I,b^\star)$, so that $\mathcal L_{\mathrm{VPD}}$ lowers the fitting error on the marked image without depending on the student's own search (Sec.~\ref{sec:evidence-quality} tests its transfer to acquired views).
In contrast, PEA's localization rollouts for $q_\theta$ and BAT's direct-answer probes and trigger context for $g_\theta$ operate exclusively on the clean, unmarked image $x=(I,q)$.
Thus, VPD teaches the answer tokens to read fine-grained evidence once localized, while PEA and BAT teach the tool-call and trigger tokens without the mark, where the evidence lies and which inputs the student fails.

Each training batch comprises 96 questions; every question contributes eight VPD answer rollouts, four PEA localization rollouts, one BAT trigger-token row, and four direct-answer probe rollouts (sampled at temperature $1.0$, as are the localization candidates) used to compute the online failure label $e_i$.
Each loss term is normalized over its respective rollout type, with weights $\lambda_{\mathrm{PEA}}=1$ and $\lambda_{\mathrm{BAT}}=0.01$.

\paragraph{Optimization.}
All parameters, including the vision encoder, are updated by AdamW (betas 0.9 and 0.999, weight decay 0.01) with one optimizer step per training step, gradient-norm clipping at 1.0, and no KL or entropy term.
The learning rate warms up over 10 steps to $2\times10^{-6}$ at 4B and $2.8\times10^{-6}$ at 9B (scaled so that the 9B weights move as much per step as the 4B weights) and then decays along a cosine schedule.
Prompts and responses are capped at 8,192 and 1,024 tokens, and all runs use seed 42 unless a seed is stated (Table~\ref{tab:seeds}).
The Qwen3.5-4B ablation runs share this recipe and change only the ablated component.

\subsection{Distillation Target and Token Weighting (VPD)}

At each valid generated token position $t\in\mathcal T$, let $\mathcal K_t$ denote the student's top-100 token IDs.
For each student or teacher distribution $p$, we form the 101-way coarsened distribution (including the residual probability mass)
\begin{equation}
\widetilde p_t=\bigl((p(a))_{a\in\mathcal K_t},\;
1-\textstyle\sum_{a\in\mathcal K_t}p(a)\bigr),
\end{equation}
clamping the residual mass for numerical stability.
We apply the contrastive reconstruction in Eq.~\eqref{eq:leap_target} to these 101-way crop and null distributions before computing JSD, with contrast weight $\beta=4$ at 4B and $\beta=8$ at 9B, and detached target parameters and top-token support.
For $M=(P+Q)/2$, the Jensen--Shannon divergence is
\begin{equation}
\mathrm{JS}_{1/2}(P,Q)=
\tfrac12\mathrm{KL}(P\|M)+\tfrac12\mathrm{KL}(Q\|M).
\end{equation}
With $\ell_t^\theta$ and $\ell_t^{\mathrm{old}}$ denoting the current and rollout actor log-probabilities of the sampled token, the detached actor-importance weight is
\begin{equation}
v_t=\operatorname{sg}\!\left[
\min\!\left(2,\exp\!\left(\operatorname{clip}
(\ell_t^\theta-\ell_t^{\mathrm{old}},-20,20)\right)\right)\right],
\end{equation}
which combines with $\omega_t$, a detached token-wise correction for the sampler/actor probability difference capped at $2$, to form $w_t=v_t\omega_t$ in Eq.~\eqref{eq:leap_vpd}.

\subsection{Localization Reward and Update (PEA)}

Region selection uses the backbone's native zoom-tool schema with normalized coordinates on $[0,1000]$, prefilling generation through the opening bracket of the box parameter.
The parser accepts up to four tool blocks, clamps finite coordinates to $[0,1000]$ before pixel conversion, and discards malformed, reversed, or empty boxes.
A response containing at least one valid box is scored by Eq.~\eqref{eq:leap_reward} using its maximum valid IoU with $b^\star$ and an extra-proposal penalty of $\eta=0.05$ per additional tool block $n_j-1$; responses with no valid box or truncation receive $r_{\mathrm{bad}}=-0.5$.
Neither reference answers nor post-crop answer accuracy enters this reward.

For a group of $M=4$ candidates per question with group-centered advantages $A_j=R_j-\frac14\sum_{k=1}^4R_k$, let the probability ratio between the current and rollout parameters $\theta$ and $\bar\theta$ at tool-call token $(j,t)$ be
\begin{equation}
\rho_{jt}=\frac{q_\theta(u_{jt}\mid x,1,u_{j,<t})}
{q_{\bar\theta}(u_{jt}\mid x,1,u_{j,<t})},
\end{equation}
with log-ratios clipped to $[-20,20]$.
Using the standard clipped surrogate
\begin{equation}
s(\rho,A)=\min\bigl(\rho A,\operatorname{clip}(\rho,0.8,1.3)A\bigr)
\end{equation}
and dual-clip bound for negative advantages
\begin{equation}
\psi(\rho,A)=\begin{cases}
\max(s(\rho,A),3A), & A<0,\\
s(\rho,A), & A\geq0,
\end{cases}
\end{equation}
PEA minimizes the GRPO-style loss~\citep{shao2024deepseekmath} over valid tool-call tokens $\mathcal U$:
\begin{equation}
\mathcal L_{\mathrm{PEA}}=-\frac1{|\mathcal U|}
\sum_{(j,t)\in\mathcal U}\omega_{jt}\psi(\rho_{jt},A_j),
\end{equation}
with unnormalized group-centered advantages, token-mean aggregation, and no entropy bonus or reference-policy KL penalty.

\subsection{Trigger Labels and Readout (BAT)}

For each question $x_i$, the soft failure target $e_i$ in Eq.~\eqref{eq:leap_bat_target} is the fraction of $K=4$ direct answers judged incorrect by the dataset's rule-based matcher.
In the inspection context, the prompt instructs the model to call the zoom tool immediately if closer inspection is needed, and otherwise to answer directly; the trigger training row evaluates the first-token probability $g_\theta(x_i)$ of \texttt{<tool\_call>} and updates $\theta$ via the centered margin surrogate in Eq.~\eqref{eq:leap_bat_loss}.
During rollouts, $g_\theta(x_i)$ is read from the top-20 next-token probabilities at 4B and exactly at 9B; the update always evaluates $g_\theta(x_i)$ exactly, and Eq.~\eqref{eq:leap_bat_loss} subtracts the batch mean explicitly, so updates remain centered at both scales.

\subsection{Inference Routing and Threshold Calibration}
\label{app:impl-inference}

At inference, BAT compares the first-token probability $g_\theta(x)$ against a fixed threshold $\tau$.
If $g_\theta(x)<\tau$, the student answers directly from $x$; if $g_\theta(x)\geq\tau$, it generates a localization response in the prefilled tool format, crops the valid predicted regions, and answers from the multi-image input $(x,Z_b)$ retaining the original image (falling back to $x$ if no valid box is produced).
The threshold $\tau$ is calibrated once on the 480 held-out examples of Vision-OPD-6K by sorting their decision scores $g_\theta(x)$ in descending order and selecting the $\lceil0.9n\rceil$-th value (targeting a 90\% inspection share on the calibration split without using answer labels), and remains fixed across the six benchmarks of Table~\ref{tab:main}.

\paragraph{Decoding.}
Every request decodes greedily with thinking disabled; images are encoded as in Vision-OPD's inference script, and the processor cap is 16,777,216 pixels.
The decision request generates one token to read $g_\theta(x)$ (from the top-20 log-probabilities at 4B and exactly from a forced one-token request at 9B).
The localization turn generates up to 512 tokens, and each box it specifies becomes a view cut from the original image.
The answer turn generates up to 16,384 tokens.

%% file: tab/supp_train_settings.tex
\begin{table}[t]
\centering
\caption{\textbf{Training settings.} The frozen teacher is a copy of the initial backbone.}
\label{tab:train-settings}
\begingroup
\tabsetup{10}{13}{6pt}
\begin{tabularx}{\linewidth}{l X X}
\toprule
Setting & LEAP-4B & LEAP-9B \\
\midrule
Backbone and teacher & Qwen3.5-4B & Qwen3.5-9B \\
Contrast weight $\beta$ & 4 & 8 \\
Peak learning rate & $2\times10^{-6}$ & $2.8\times10^{-6}$ \\
\midrule
\tabgroup{3}{Shared}
Training data & \multicolumn{2}{l}{Vision-OPD-6K} \\
Questions per step & \multicolumn{2}{l}{96, one optimizer step} \\
Rows per question & \multicolumn{2}{l}{8 VPD, 4 PEA, 4 probes, 1 BAT} \\
Sampling temperature & \multicolumn{2}{l}{1.0} \\
Prompt / response limit & \multicolumn{2}{l}{8,192 / 1,024 tokens} \\
Optimizer & \multicolumn{2}{l}{AdamW (0.9, 0.999), weight decay 0.01} \\
Schedule & \multicolumn{2}{l}{10 warm-up steps, then cosine decay} \\
Gradient clipping & \multicolumn{2}{l}{1.0} \\
Loss weights & \multicolumn{2}{l}{$\lambda_{\mathrm{PEA}}=1$, $\lambda_{\mathrm{BAT}}=0.01$} \\
Trained parameters & \multicolumn{2}{l}{all, vision encoder included} \\
Seed & \multicolumn{2}{l}{42} \\
\bottomrule
\end{tabularx}
\endgroup
\end{table}

%% file: sec/E_intervals.tex
\section{Paired Differences and Intervals}
\label{app:intervals}

This section reports 95\% bootstrap intervals for the core comparisons behind the main claims of Secs.~\ref{sec:introduction}, \ref{sec:leap}, and~\ref{sec:experiments}.
The intervals are percentile intervals from paired bootstrap resampling over the evaluation questions answered by both compared conditions (2,000 resamples).
HR-Bench is resampled by question cluster, inspection AUROC by image cluster within each benchmark, and HallusionBench readouts in Table~\ref{tab:bias} by figure family; the error bars of the AUROC curves and the ZoomBench accuracies in Fig.~\ref{fig:gain-source}b and~c come from the same resampling.
Each paragraph below covers one part of the main text, and each entry gives the point estimate (in percentage points or AUROC) followed by its 95\% interval.

\paragraph{Imitation-gap diagnosis (Secs.~\ref{sec:introduction} and~\ref{sec:leap_gap}).}
For end-to-end yes-bias in our controlled Qwen3.5-4B distillation, HallusionBench false affirmation rises over the backbone by $+8.6$ $[5.0,12.3]$ points under crop-teacher distillation and by $+12.5$ $[9.2,15.8]$ points under VPD.
Across the broader distillation family (Sec.~\ref{sec:leap_gap}, POPE resampled by image), false alarms increase over the same-size Qwen3.5 base by $+9.7$ $[8.5,11.0]$, $+9.3$ $[8.1,10.5]$, $+7.7$ $[6.6,8.8]$, and $+14.3$ $[12.9,15.8]$ points for Vision-OPD-4B, VAD-4B, OPD-Aha-4B, and OPD-V-4B, and by $+1.4$ $[0.7,2.2]$, $+5.1$ $[4.1,6.2]$, $+10.5$ $[9.3,11.9]$, and $+0.8$ $[0.1,1.5]$ points for the corresponding 9B students.
For behavioral tool-call suppression, native zoom calls shift from the backbone's 100\% by $-51.7$ $[-55.4,-48.1]$ points under crop-teacher distillation and by $-87.0$ $[-88.8,-85.2]$ points under VPD, while on V$^\star$ and ZoomBench RS-OPSD-8B shifts native zoom calls by $-45.8$ $[-49.3,-42.1]$ points relative to Qwen3-VL-8B.
The inspection-decision error-ranking AUROC is 0.56 $[0.54,0.59]$ for the backbone, 0.44 $[0.42,0.47]$ for the crop teacher, 0.39 $[0.37,0.42]$ for VPD (already dropping to 0.45 $[0.43,0.48]$ by step 20), and 0.44 $[0.41,0.46]$ for the released Vision-OPD-4B student.

\paragraph{Main benchmark comparisons and inference controls (Sec.~\ref{sec:main-results}).}
On Avg$_6$ at 4B, LEAP exceeds Vision-OPD by $+5.50$ $[4.14,6.88]$ points and OPD-Aha by $+1.71$ $[0.60,2.84]$ points over their common valid items.
Across the four high-resolution perception benchmarks (V$^\star$, ZoomBench, and HR-Bench 4K/8K), LEAP-4B exceeds Vision-OPD-4B and VAD-4B under zooming by $+14.01$ $[10.69,17.30]$ and $+7.82$ $[4.57,11.23]$ points and exceeds a three-sample majority vote over LEAP-4B's single pass at a matched token budget ($33.0$ thousand tokens per question against $33.3$ thousand for LEAP-4B) by $+2.81$ $[1.38,4.34]$ points (Table~\ref{tab:tool-loops}).
Over its own single pass on these benchmarks, zooming raises LEAP-4B by $+3.01$ $[1.55,4.47]$ points.

\paragraph{Component and target ablations (Sec.~\ref{sec:ablations}).}
For answer distillation and acquisition complementarity, full LEAP improves over VPD by $+3.31$ $[0.83,5.80]$ ZoomBench points, with a positive $+2.18$ $[0.67,3.61]$-point interaction between target contrast and acquisition training in the gain from acquired views; replacing the null view with a degraded crop changes Avg$_6$ by $-1.35$ $[-2.26,-0.48]$ points.
For where to look, PEA raises the ZoomBench evidence hit rate by $+4.06$ $[1.29,6.85]$ points with VPD and BAT fixed and by $+5.75$ $[2.86,8.60]$ points over VPD without BAT, and replacing its overlap reward with an outcome reward changes Avg$_6$ by $-2.00$ $[-3.06,-0.97]$ points.
For when to look, VPD + PEA remains inverted at 0.38 $[0.35,0.40]$ AUROC; adding BAT to it raises AUROC to 0.62 $[0.59,0.64]$ for LEAP-4B, a paired gain of $+0.24$ $[0.20,0.28]$, whereas adding BAT to VPD alone gives 0.57 $[0.54,0.60]$, and LEAP-9B reaches 0.62 $[0.60,0.65]$.
In native zoom invocation, BAT and PEA exhibit a $+65.1$ $[62.1,68.3]$-point positive interaction.

\paragraph{Mitigating the imitation gap (Sec.~\ref{sec:evidence-quality}).}
On ZoomBench, replacing LEAP's self-acquired views with random-box crops or noise images lowers accuracy by $7.93$ $[4.85,10.77]$ and $7.57$ $[4.73,10.53]$ points, and on the 110 items where acquired views miss the evidence, substituting GT crops raises accuracy by $+13.64$ $[6.36,21.82]$ points.
For native tool-use behavior (Table~\ref{tab:native}), deciding autonomously when to zoom lowers V$^\star$/ZoomBench accuracy by $6.28$ $[3.26,9.56]$, $2.37$ $[0.46,4.44]$, and $7.09$ $[4.55,9.79]$ points for Vision-OPD, OPD-Aha, and OPD-V.

\paragraph{Remote-sensing transfer (Sec.~\ref{sec:remote-sensing}).}
Under the shared corrected judge and stratified question resampling, zero-shot LEAP-9B exceeds the domain-trained RS-OPSD-8B on XLRS-Bench-lite by $+2.82$ $[0.14,5.52]$ points and on MME-RealWorld-RS by $+3.68$ $[2.09,5.21]$ points and differs from it on LRS-VQA by $-3.32$ $[-4.68,-1.98]$ points.
With in-domain remote-sensing training, our 4B model LEAP-RS exceeds RS-OPSD-8B overall by $+2.15$ $[1.03,3.25]$ average points.

%% file: sec/G_prompts.tex
\section{Prompt Templates}
\label{app:prompts}

The boxes below give the prompts used in evaluation, training, and inference.
Prompt text is set in typewriter type and placeholders in italics; \ph{query} is a benchmark question with its options and answer instruction, and line breaks are omitted except in the judge prompt and the localization prefill.
Direct-answer probes during training and direct answers at inference send \ph{query} with the clean image and no other text.

\begin{promptbox}{Benchmark answer instructions (Appendix~\ref{app:benchmarks})}
\textbf{V$^{\star}$:} options (A), (B), \ldots, then\\
\texttt{Answer with the option\textquotesingle s letter from the given choices directly.}\\[2pt]
\textbf{HR-Bench:} \texttt{Select from the following choices.} and options (A) to (D).\\[2pt]
\textbf{MME-RealWorld:} \texttt{The choices are listed below:} and options (A) to (E), then\\
\texttt{Select the best answer to the above multiple-choice question based on the image. Respond with only the letter (A, B, C, D, or E) of the correct option. The best answer is:}\\
(translated into Chinese for the Chinese split).\\[2pt]
\textbf{ZoomBench:} released questions; counting questions end with \texttt{Please answer using Arabic numerals.}
\end{promptbox}

\begin{promptbox}{Judge prompt (Vision-OPD, gpt-oss-120b; Appendix~\ref{app:eval-protocol})}
\ttfamily
Your task is to judge whether the response expresses the same meaning as the answer of a question.\\
The question is: \ph{question}\\
The answer is: \ph{gold answer}\\
The response is: \ph{extracted response}\\
Please check and compare them and then judge. If the response is correct, your output should be Yes. Otherwise, your output should be No. Directly give me your output.
\end{promptbox}

\begin{promptbox}{VPD rows (Vision-OPD-6K prompt)}
\ph{question} \texttt{Only focus on the objects inside the red bounding box in the image to answer this question.} \ph{options}\\[2pt]
The student receives this prompt with the region-marked full image, and the teacher with the crop.
\end{promptbox}

\begin{promptbox}{Inspection decision, trigger rows, and localization rollouts}
\textbf{System:} Qwen-Agent's \texttt{image\_zoom\_in\_tool} declared in Qwen3.5's native tool format, with a box \texttt{bbox\_2d} $[x_1,y_1,x_2,y_2]$ on a 0--1000 grid, a \texttt{label}, and an \texttt{img\_idx}.\\[2pt]
\textbf{User:} the clean image, then\\
\ph{query} \texttt{If you need to look closer, call image\_zoom\_in\_tool immediately as your first output, with no text before it. Otherwise answer the question directly.}
\end{promptbox}

\begin{promptbox}{Localization prefill}
\texttt{<tool\_call>}\\
\texttt{<function=image\_zoom\_in\_tool>}\\
\texttt{<parameter=bbox\_2d>}\\
\texttt{[}\\[2pt]
Localization continues the assistant turn from this prefill, which ends inside the box parameter.
\end{promptbox}

\begin{promptbox}{Answer turn after inspection}
\textbf{User:} the full image, then\\
\texttt{Here are the regions you selected, cropped from the original full-resolution image and enlarged, in the order you listed them. Use them together with the full image to answer.}\\
then for each view $k$\\
\texttt{View $k$: bbox\_2d $[x_1,y_1,x_2,y_2]$ of the full image, its \ph{position} part:}\\
followed by the view, and finally \ph{query}.
\end{promptbox}

%% file: sec/H_case_studies.tex
\section{Case Studies}
\label{app:cases}

Figs.~\ref{fig:cases-zoombench} and~\ref{fig:cases-more} show eight questions from ZoomBench, HR-Bench 4K, and MME-RealWorld that LEAP-4B answers correctly from an acquired view covering 0.079\% to 5.157\% of the image.
Vision-OPD-4B and OPD-Aha-4B answer from the full image in a single pass and err on all eight.
SenseNova-MARS crops the queried object with its tool loop in every example, yet answers only one of the eight correctly.
The three closed-source models answer four or five correctly from the full image; on the tower clock all six compared models return 11:40, and none counts the four boards in the remote-sensing image.

\input{fig/figS_case.tex}

%% file: fig/figS_case.tex
\begin{landscape}
\begin{figure}[p]
\centering
\begin{minipage}{0.84\linewidth}
\includegraphics[width=\linewidth,page=1]{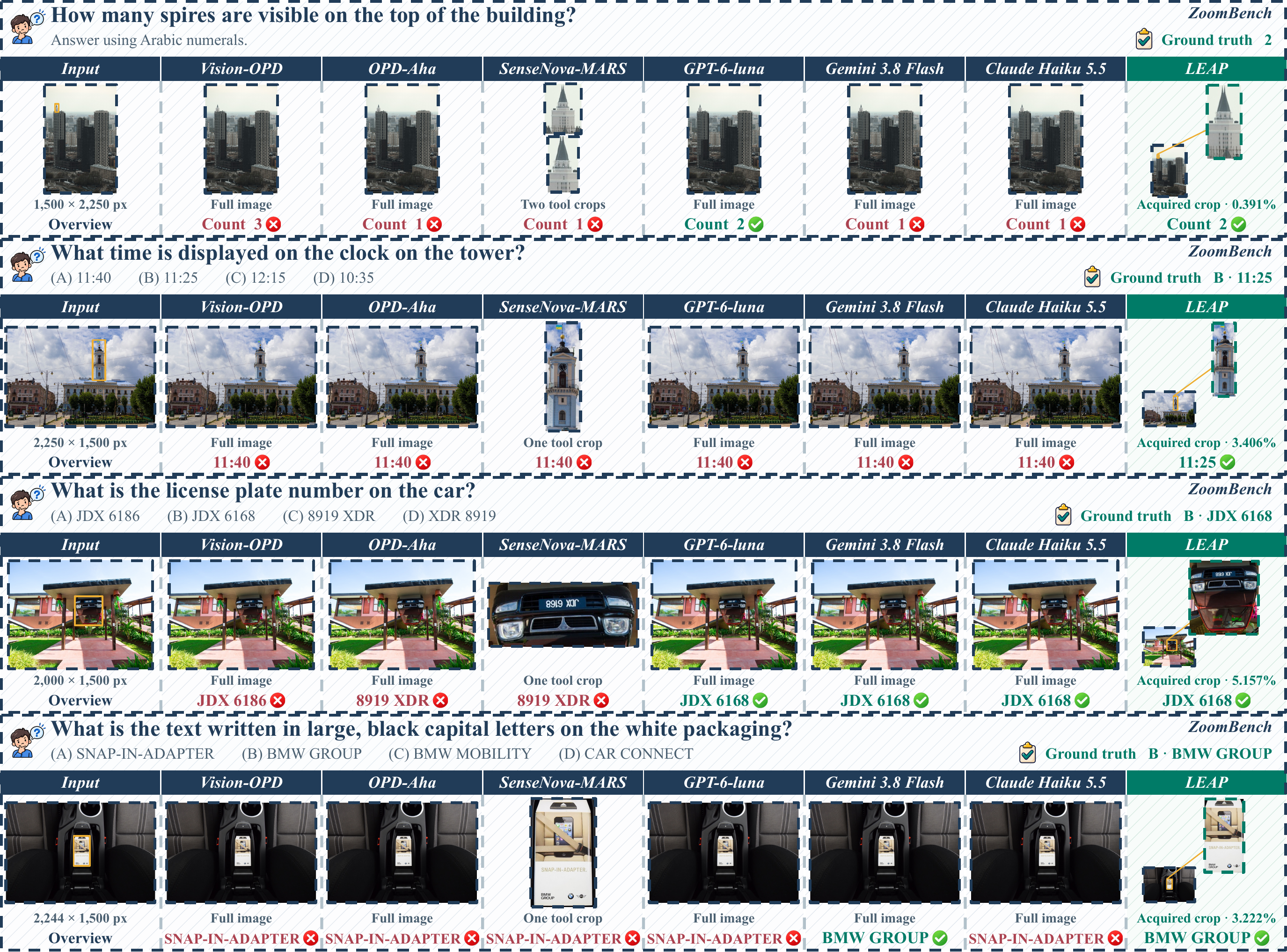}
\caption{\textbf{Case studies on ZoomBench.}}
\label{fig:cases-zoombench}
\end{minipage}
\end{figure}
\begin{figure}[p]
\centering
\begin{minipage}{0.84\linewidth}
\includegraphics[width=\linewidth,page=2]{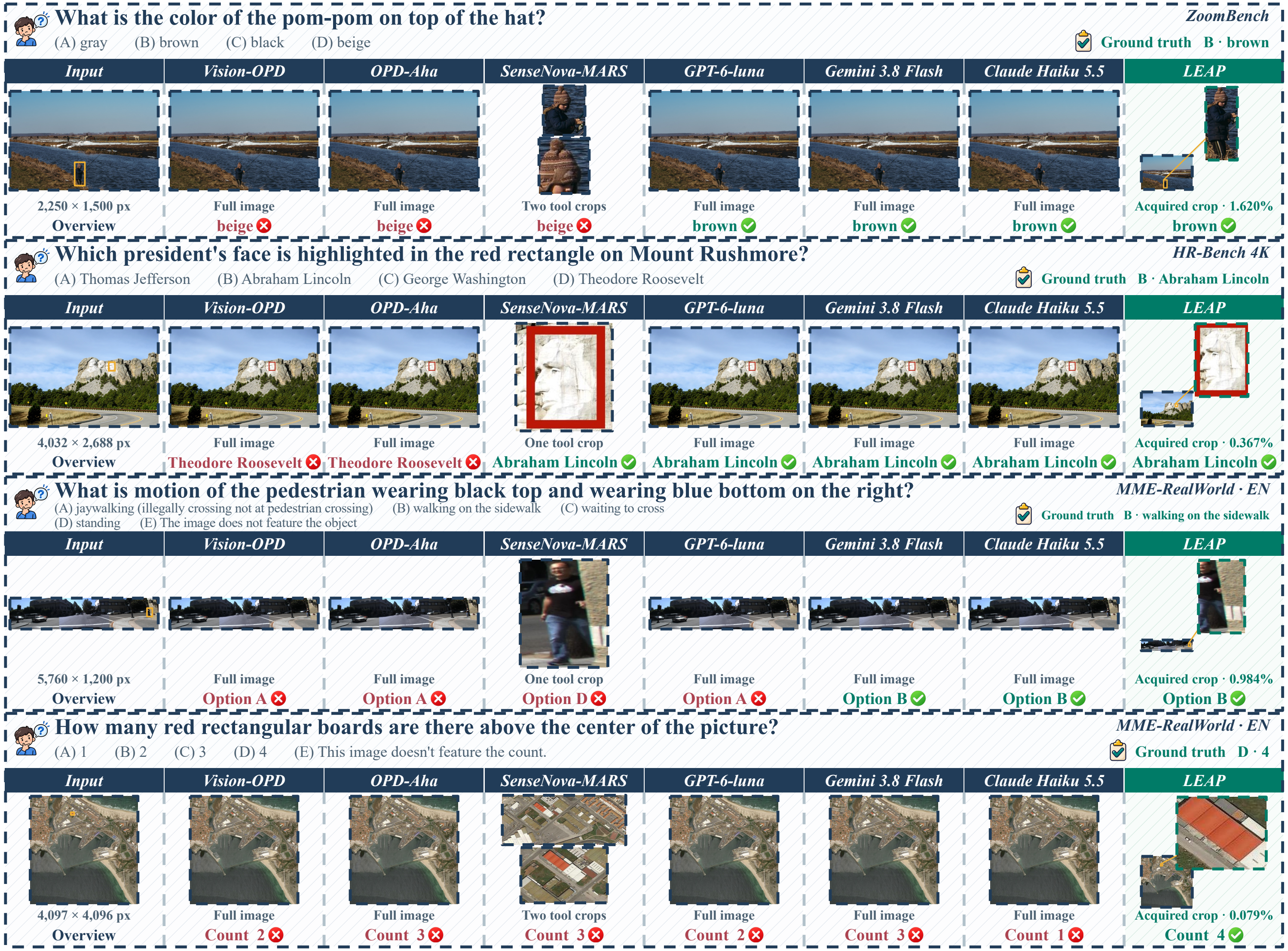}
\caption{\textbf{Case studies on ZoomBench, HR-Bench 4K, and MME-RealWorld.}}
\label{fig:cases-more}
\end{minipage}
\end{figure}
\end{landscape}